\documentclass[10pt,journal,compsoc]{IEEEtran}
\usepackage{graphicx}
\usepackage{subcaption}
\usepackage{enumitem}
\usepackage{hyperref}
\hypersetup{colorlinks=true,linkcolor=blue,anchorcolor=blue,citecolor=blue}
\usepackage{multirow}
\usepackage{setspace}
\usepackage{multicol}
\usepackage{stfloats}

\usepackage[table]{xcolor}
\definecolor{lightgray}{RGB}{220, 220, 220}
\usepackage{xspace}
\usepackage{bbding}
\usepackage{pifont}
\usepackage{ragged2e}
 \usepackage{makecell}
\newcommand{\etal}{\emph{et al.}\xspace}

\newcommand{\myblue}[1]{{\color{blue}#1}}
\definecolor{mygreen_rgb}{RGB}{0,0,0}

\usepackage{algorithm}
\usepackage{algorithmic}
\usepackage{amsmath}

\usepackage{amssymb}
\usepackage{booktabs}
\usepackage{multirow}

\begin{document}

\title{\huge EgoLoc: A Generalizable Solution for Temporal Interaction Localization in Egocentric Videos}%

\author{Junyi~Ma$^{*}$, Erhang~Zhang$^{*}$, Yin-Dong Zheng, Yuchen Xie, Yixuan Zhou, Hesheng~Wang$^{\dag}$
\thanks{This work was supported by National Key R\&D Program of China (Grant No.2024YFB4708900). It was also supported in part by the Natural Science Foundation of China under Grants 62225309, U24A20278, 62361166632, and U21A20480.}
\thanks{Junyi~Ma, Erhang~Zhang, Yin-Dong~Zheng, Yuchen Xie, Yixuan~Zhou, and Hesheng~Wang are with IRMV Lab, the Department of Automation, Shanghai Jiao Tong University, Shanghai 200240, China.}
\thanks{$^{*}$Equal contribution}
\thanks{$^{\dag}$Corresponding author email: wanghesheng@sjtu.edu.cn}
}

\IEEEtitleabstractindextext{
\begin{abstract}
Analyzing hand-object interaction in egocentric vision facilitates VR/AR applications and human-robot policy transfer. Existing research has mostly focused on modeling the behavior paradigm of interactive actions (i.e., ``how to interact''). However, the more challenging and fine-grained problem of capturing the critical moments of contact and separation between the hand and the target object (i.e., ``when to interact'') is still underexplored, which is crucial for immersive interactive experiences in mixed reality and robotic motion planning. Therefore, we formulate this problem as \textbf{temporal interaction localization (TIL)}. Some recent works extract semantic masks as TIL references, but suffer from inaccurate object grounding and cluttered scenarios. Although current temporal action localization (TAL) methods perform well in detecting verb-noun action segments, they rely on category annotations during training and exhibit limited precision in localizing hand-object contact/separation moments. To address these issues, we propose a novel zero-shot approach dubbed \textit{EgoLoc} to localize hand-object contact and separation timestamps in egocentric videos. EgoLoc introduces hand-dynamics-guided sampling to generate high-quality visual prompts. It exploits the vision-language model to identify contact/separation attributes, localize specific timestamps, and provide closed-loop feedback for further refinement. EgoLoc eliminates the need for object masks and verb-noun taxonomies, leading to generalizable zero-shot implementation. Comprehensive experiments on the public dataset and our novel benchmarks demonstrate that EgoLoc achieves plausible TIL for egocentric videos. It is also validated to effectively facilitate multiple downstream applications in egocentric vision and robotic manipulation tasks. Code and relevant data will be released at \url{https://github.com/IRMVLab/EgoLoc}. 
\end{abstract}


}

\maketitle
\IEEEdisplaynontitleabstractindextext
\IEEEpeerreviewmaketitle

\section{Introduction}
\label{sec:intro}

\begin{figure}[t]
  \centering
  \captionsetup{aboveskip=2pt, belowskip=0pt}
  \includegraphics[width=0.95\linewidth]{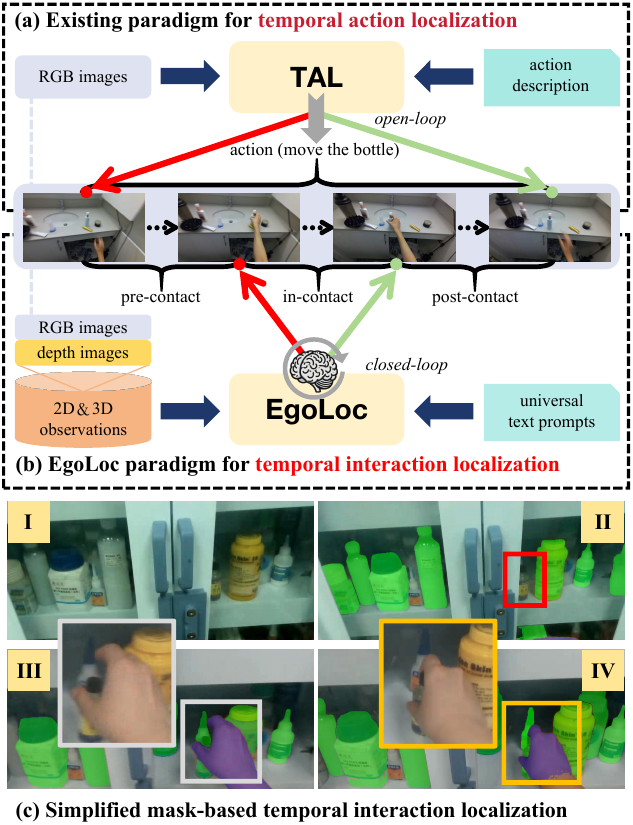}
  \caption{Compared to the TAL paradigm (a) that localizes a designated action in an untrimmed video with RGB images and action descriptions, TIL aims to find the frames closest to the start and end of a human-object in-contact period. Our proposed EgoLoc paradigm (b) absorbs both 2D and 3D observations and implements accurate TIL with universal text prompts in a closed-loop form. Here we also provide an example of mask-based TIL (c), where (c)-I denotes the interaction scene and (c)-II/III/IV illustrate the HOI process with hand masks (blue) and object masks (green) generated by Grounded SAM. EgoLoc avoids the inherent flaws of the ad-hoc masked-based method, showing promising potential for downstream applications.}  
  \label{fig:motivation}
  \vspace{-0.5cm}
\end{figure}

\IEEEPARstart{H}{and}-object interaction (HOI) analysis has witnessed significant advancements in the field of egocentric vision~\cite{betancourt2015evolution,plizzari2024outlook,bandini2020analysis}, enabling various downstream VR/AR and robotic applications. Despite significant progress over recent years in investigating HOI reconstruction and synthesis \cite{on2025bigs,liu2024easyhoi,Ye_2023_ICCV,zhang2024hoidiffusion,zhou2024gears}, hand mesh recovery~\cite{pavlakos2024reconstructing,tu2023consistent,ren2024pyramid,Li_2024_CVPR,tian2023recovering}, and interaction region extraction~\cite{zhang2022fine,li2022egocentric,wang2025precision,roy2024interaction,fan2021understanding}, the aims of HOI research are primarily limited in better understanding ``how to interact'' in the spatial domain. However, when it comes to complex downstream applications such as immersive interaction in mixed reality and robot action planning in pick-and-place tasks, capturing ``when to interact'' in the temporal domain becomes inevitable. Concretely, localizing the timestamps of hand-object contact and separation for first-person videos should be further explored in the egocentric vision literature. To this end, we formulate this temporal localization problem as \textbf{temporal interaction localization (TIL)}, where we regard any HOI process as three distinct stages: pre-contact, in-contact, and post-contact. As a companion study, temporal action localization (TAL)~\cite{zeng2021graph,gao2023pami,bao2022opental,li2024detal,xu2025transferable,abdullah2025ual}, which determines the time durations of specific actions from long untrimmed videos (Fig.~\ref{fig:motivation}(a)), has been widely studied. In contrast, TIL aims to pinpoint the egocentric frames closest to the start and end of a hand-object in-contact period, accurately distinguish the three stages for each HOI process (Fig.~\ref{fig:motivation}(b)).
The fundamental distinction between TIL and TAL stems from their respective granularity requirement. TAL captures coarse action transitions while TIL attends to fine-grained contact/separation timestamps.

Nevertheless, transferring TAL methods to TIL in the context of egocentric videos could serve as a reasonable starting point, while the following challenges inevitably hinder such adaptation:

\begin{itemize}[leftmargin=1em]
\item \textbf{Granularity}: Current TAL methods are primarily designed to detect action or task transitions without explicitly modeling the HOI process. While these methods effectively identify macroscopic changes in movement patterns, they struggle to precisely distinguish interaction states between the hand and the object at the microscopic level.
Consequently, they lack the capability to accurately find the exact timestamps of hand-object contact and separation, which is required by TIL tasks.
\item \textbf{Generalization capability}: Most TAL models necessitate optimization with labor-intensive annotations. Thus, they cannot generalize to out-of-distribution HOI categories. While recent zero-shot temporal action localization (ZS-TAL)~\cite{wake2024open,gupta2024open} uses large vision-language models (VLMs) for better generalizability, its performance remains constrained by predefined action categories in text prompts. 
\item \textbf{Perception capability}: 
TAL methods typically take as input 2D RGB images, while failing to accommodate 3D spatial information. However, HOI occurs in 3D space, where 3D hand dynamics may provide richer interaction signals with an absolute scale to enhance HOI understanding.
\item \textbf{Closed-loop capability}: TAL methods generally operate in an open-loop manner, where any slight timing deviation cannot be corrected using a posteriori information. The resulting error also tends to be amplified in the final output.
This limitation leads to high estimation uncertainty when adapting to TIL, as the localization targets in TIL are much finer-grained than those in TAL.
\end{itemize}

In recent works for human-robot skill transfer~\cite{ren2025motion,chen2024vlmimic,wang2024vlm}, a straightforward yet popular ad-hoc solution is extracting semantic masks of hands and target objects (see Fig.~\ref{fig:motivation}(c)) to measure the spatial relationship between them. TIL can be coarsely operated by comparing hand-object distances with preset thresholds or quantifying hand keypoint occupancy within the object mask. However, such mask-based temporal interaction localization suffers from several inherent flaws:
\begin{itemize}[leftmargin=1em]
\item \textbf{Inaccurate object mask:} In most HOI cases, it is difficult to capture highly accurate descriptions of target object instances. Therefore, only vague text prompts can be used by visual grounding models, leading to inaccurate masks for specific target instances. For example, in the HOI scene of Fig.~\ref{fig:motivation}(c)-I, Grounded SAM~\cite{ren2024grounded} prompted by \textit{bottle} generates masks for both target and non-target bottles in Fig.~\ref{fig:motivation}(c)-II/III/IV. In addition, some target objects have ``long-tail" shapes, which are hard to identify with visual grounding models. For instance, the target bottle within the red bounding box in Fig.~\ref{fig:motivation}(c)-II sometimes cannot be detected by Grounded SAM given \textit{bottle}.

\item \textbf{Scale aliasing from 2D observation:} 2D semantic relationship may be inconsistent with actual 3D HOI styles. Therefore, the foreground and background distractors affect interaction accuracy. As shown in Fig.~\ref{fig:motivation}(c)-III, the thumb tip and the index fingertip present a high probability of contact with the bottle behind the target instance, upon only considering the 2D relationship between hand keypoints and object masks. Besides, the hand and target instance oftentimes overlap in the 2D image plane despite being spatially distant in 3D space, leading to incorrect interaction localization. This also stems from the scale aliasing of 2D observations, which cannot describe accurate 3D structural HOI information.

\item \textbf{Occluded target object:} During the interaction process in cluttered scenes, the target objects are often occluded by hands or other objects (see Fig.~\ref{fig:motivation}(c)-IV). In these cases, it is hard to extract an accurate hand-object relationship due to incomplete semantic masks of target objects. Borrowing the object mask from other frames without occlusion is also infeasible since the object mask moves and deforms among different image planes due to camera egomotion.
\end{itemize}

Therefore, a dedicated paradigm should be specially designed for more reasonable TIL compared to the above-mentioned TAL approaches and mask-based schemes. 
To this end, we propose \textbf{EgoLoc, a novel zero-shot temporal interaction localization (ZS-TIL) paradigm for egocentric videos}.
EgoLoc localizes the timestamps at which a human hand makes contact with and separates from an object in a long untrimmed video. 
We attend to diverse hand-object interactions that evoke grasping actions, such as picking and placing, pouring, hinge opening and closing, slide opening and closing, cutting, and peeling.
EgoLoc has enhanced perception capability because it parses both 2D RGB images and 3D point clouds to extract 3D hand dynamics. This facilitates a devised self-adaptive sampling strategy, which generates high-quality initial guesses of the frames around the start (contact) and end (separation) of HOI. The following VLM-based localization module first exploits the VLM discriminator to identify contact/separation attributes of the sampled initial guess, and outputs first-round contact/separation timestamps with the VLM localizer. Ultimately, the closed-loop feedback mechanism introduces the VLM checker to measure the plausibility of the first-round localization results. The accepted first-round results can be directly output, while the rejected ones facilitate the second-round reasoning with additional in-context learning. 

As can be noted in Fig.~\ref{fig:motivation}(a) and Fig.~\ref{fig:motivation}(b), compared to the existing TAL methods, EgoLoc localizes finer-grained timestamps of HOI in a closed-loop manner. It has a stronger generalization capability thanks to the powerful VLM using action-category-agnostic prompts. Additionally, EgoLoc captures richer interaction information by accommodating sequential 2D and 3D observations. Compared to the mask-based schemes in Fig.~\ref{fig:motivation}(c), EgoLoc does not require explicit target object masks, and thus is not affected by their grounded quality. The scale aliasing problem is also mitigated by introducing 3D observations and taking as input the image sequence for VLM reasoning. The main contributions of this work are as follows:

\begin{itemize}[leftmargin=1em]
    \item \textbf{Novel TIL paradigm:} We propose the first zero-shot temporal interaction localization paradigm, namely EgoLoc, for egocentric videos. It broadens the scope of egocentric vision research.
    \item \textbf{Hand dynamics incorporation:} To the best of our knowledge, EgoLoc is the first work to incorporate 3D hand dynamics into temporal localization. A novel self-adaptive sampling strategy is designed based on hand motion analysis to generate high-quality initial guesses for VLM-based TIL reasoning. 
    \item \textbf{Closed-loop feedback:} We introduce a feedback mechanism to check and refine the TIL results. It closes the inference loop to reduce estimation uncertainties and improve localization accuracy.
    \item \textbf{New benchmarks and extensive experiments:} We build novel benchmarks, DeskTIL and ManiTIL, to facilitate future TIL research. Comprehensive experiments on the public dataset, our newly proposed benchmarks, and multiple downstream tasks demonstrate that EgoLoc achieves plausible temporal interaction localization, with significant potential for supporting downstream egocentric vision and robotic manipulation applications.
\end{itemize}

This work is a substantial extension of our preliminary version~\cite{zhang2025egoloc}, which introduces new elements as follows:

\begin{itemize}[leftmargin=1em]
\item Relax the assumption that only one single HOI process exists in each video.
\item Relax the assumption that the contact moment must occur in the first half of each video.
\item Refine the devised components in EgoLoc including: (1) extending hand motion analysis by incorporating both velocities and accelerations for the self-adaptive sampling strategy, (2) introducing hand grounding into the VLM-based localization module, and (3) streamlining the closed-loop feedback mechanism by removing ad-hoc speed-based checks and introducing in-context learning. 

\item Propose a novel benchmark, ManiTIL, to conduct more comprehensive evaluation and facilitate future works along with the prior DeskTIL benchmark.
\item Validate EgoLoc's practicality on multiple downstream egocentric vision and robotic manipulation tasks. 
\end{itemize}

The rest of this article is organized as follows. Sec.~\ref{sec:related} reviews the related works in TAL and HOI analysis in egocentric vision. Sec.~\ref{sec:method} details the proposed EgoLoc paradigm. Experimental setups, results, and analysis are reported in Sec.~\ref{sec:experiments}. Ultimately, Sec.~\ref{sec:conclusion} summarizes this work.

\section{Related Works}
\label{sec:related}

\subsection{Temporal Action Localization}

Recent years have witnessed significant advancements in temporal action localization (TAL)~\cite{zeng2019graph,phan2024zeetad,li2024detal,yan2023unloc,vahdani2022deep}. While conventional action recognition techniques~\cite{zheng2020dynamic,bao2021evidential,Shiota_2024_WACV,chi2024infogcn,peng2024referring} classify action categories of video clips, TAL focuses on identifying both the initiation and termination timestamps of action instances within continuous image streams. Thus, TAL can be regarded as a companion study with respect to temporal interaction localization (TIL) in this work. Since our method is developed based on VLMs to achieve zero-shot inference, here we only review the emerging VLM-based TAL approaches. We refer more traditional counterparts to the survey by Wang~\etal~\cite{wang2023temporal}. 

As a pioneering work that introduces VLM into the TAL paradigm, Ju~\etal~\cite{ju2022prompting} add learnable prompt vectors for CLIP~\cite{radford2021learning} inputs and use a lightweight Transformer for temporal feature modeling. Similar to this work, Nag~\etal~\cite{nag2022zero} propose STALE, which also uses a standard CLIP pre-trained Transformer with learnable prompts, but additionally introduces a novel class-agnostic representation masking concept to enable zero-shot action detection. To address the problem of lacking temporal relations and motion cues in STALE's frame
features, Phan~\etal~\cite{phan2024zeetad} further associate VLM-encoded semantics with dynamic foreground masks. Considering weak vision-language interactions of these approaches, Raza~\etal~\cite{raza2024zero} bridge text and vision branches by pre-training multimodal
prompts on a recognition dataset, leading to better zero-shot reasoning.
More recently, T3AL~\cite{liberatori2024test} also performs zero-shot TAL by test-time adaptation, which relaxes the requirement for training data. In the above works, VLMs typically serve as a feature extractor for input video clips. In contrast, some other works attend to the advanced reasoning skills of large language models (LLMs) and VLMs for zero-shot and open-vocabulary TAL. For instance, Ju~\etal~\cite{ju2023multi} propose using action-specific textual descriptions from LLMs and visual-contextualized instance-wise prompt embeddings, endowing classifiers with strong discriminative power.
Gupta~\etal~\cite{gupta2024open} prompt LLMs to generate rich class-specific language descriptions for image streams, guiding visual cues and semantic context learning. Some VLM-based moment retrieval methods~\cite{li2024mvbench,luo2024zero} can also be adapted to TAL with the capability of long video understanding.
Recent work T-PIVOT~\cite{wake2024open} proposed by Wake~\etal directly leverages GPT-4o~\cite{achiam2023gpt} to infer action timings by analyzing tiled image sequences extracted from long videos. Building upon PIVOT~\cite{nasiriany2024pivot}, it uses an iterative window refinement strategy to progressively localize key frames. In this work, we also develop our TIL paradigm with the off-the-shelf VLM to improve the zero-shot reasoning ability. In contrast to the above TAL paradigms that only capture coarse action transitions, our proposed EgoLoc extracts finer-grained timings of human-object contact/separation states in the task of temporal interaction localization for egocentric videos. Moreover, we introduce additional 3D perceptions to enhance spatial understanding, compared to the existing TAL methods that only process 2D image streams. Considering high uncertainties inherent in open-loop localization schemes, we also develop a closed-loop mechanism to further refine the VLM-based localization results.

\subsection{HOI Analysis in Egocentric Vision}

Comprehending how a hand interacts with an object has been widely investigated in the literature of egocentric vision~\cite{betancourt2015evolution,plizzari2024outlook}. The first-person camera observations have human-like viewpoints, offering visual cues for HOI analysis. In earlier days, representing HOI with detected hand joints and object bounding boxes~\cite{fan2018forecasting,damen2018scaling,shan2020understanding} has limited practical applicability for downstream tasks since it cannot fully capture specific hand-object shapes. Therefore, many following HOI reconstruction methods~\cite{chen2021joint,wang2023interacting,on2025bigs} switch to hand-object pose estimation and mesh recovery. More recently, HOI synthesis techniques~\cite{zhang2024hoidiffusion,ye2023affordance,xue2024hoi} have been proposed to directly synthesize hand-object meshes or realistic images with visual or textual prompts. Instead of concurrently comprehending hand and object states, some object-agnostic HOI analysis techniques have been proposed in fields such as hand mesh recovery~\cite{pavlakos2024reconstructing,tian2023recovering,prakash20243d} and hand motion forecasting~\cite{bao2023uncertainty,ma2024diff,ma2024madiff}, which do not consider object categories and shapes. With the rapid development of LLMs and VLMs, extensive works introduce their strong generalizability into HOI analysis frameworks. For instance, Mittal~\etal~\cite{mittal2024can} integrate Q-former~\cite{li2023blip} with an LLM to construct a video-language model for egocentric HOI action anticipation. In contrast, Rai \etal~\cite{rai2024strategies} utilize knowledge from VLMs to accomplish weakly supervised affordance grounding. MADiff~\cite{ma2024madiff} exploits vision-language features from a foundation model for semantic scene understanding.
Egothink~\cite{cheng2024egothink} and EgoVLM~\cite{vinod2025egovlm} both present the robust ability of multiple advanced VLMs to reason about HOI contents from first-person videos. Compared to these HOI analysis techniques focusing on ``how hands interact with objects'', our EgoLoc attends to egocentric temporal interaction localization, enlarging the domain of HOI analysis by \textbf{considering ``when hands meet objects''}. EgoLoc is also agnostic to explicit target object masks, which alleviates the flaws inherent in conventional mask-based solutions. It exploits hand dynamics and contextual visual cues, showing strong TIL performance across multiple cluttered scenes.

\section{Methodology}
\label{sec:method}

\begin{figure*}
  \centering
  \captionsetup{aboveskip=2pt, belowskip=0pt}
  \includegraphics[width=1\linewidth]{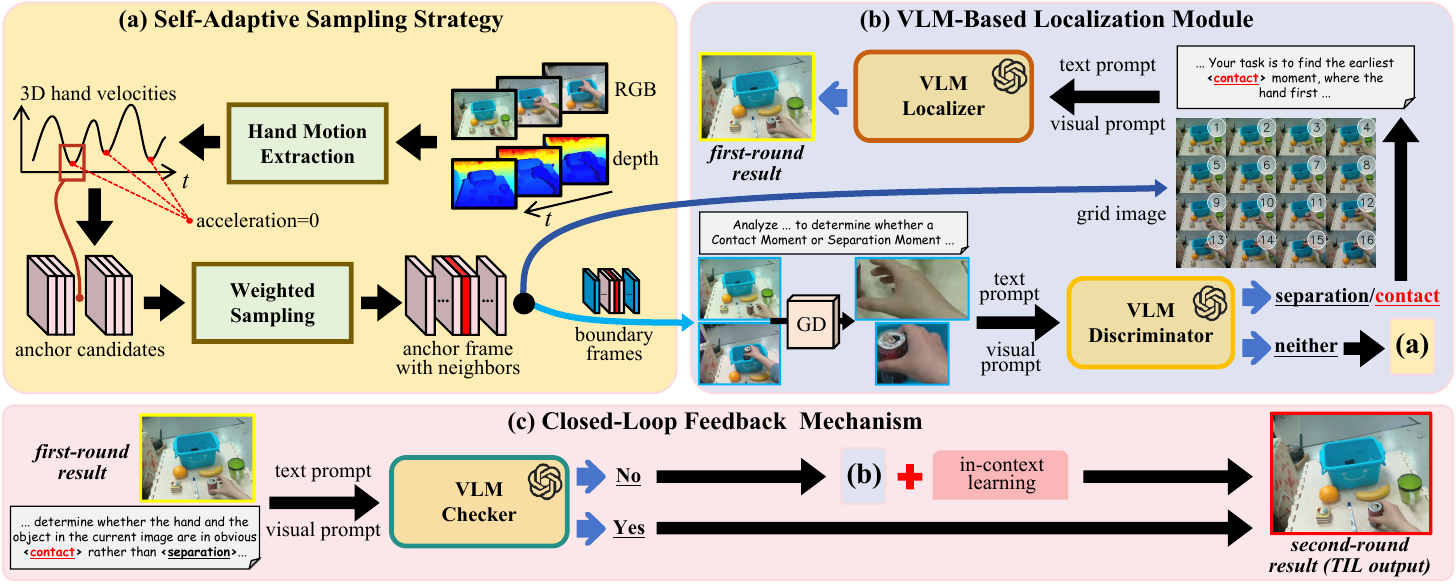}
  \caption{Overall pipeline of EgoLoc. EgoLoc integrates the self-adaptive sampling strategy (a), the VLM-based localization module (b), and the closed-loop feedback mechanism (c). Firstly, the self-adaptive sampling strategy samples the anchor frame of interaction transition timestamps based on 3D hand dynamics extracted from RGB and depth images. Then, the VLM-based localization module identifies the contact/separation attributes and localizes the first-round timestamp. Ultimately, the closed-loop feedback mechanism outputs the refined second-round timestamp by assessing hand-object visual cues and in-context learning.}
  \label{fig:pipeline}
  \vspace{-0.49cm}
\end{figure*}

\subsection{Task Definition}
\label{sec:task_def}
In this section, we first provide the detailed definition of the temporal localization task.
Given a long untrimmed egocentric video with $N_\text{obs}$ sequential RGB images $\mathcal{I}=\{I_t\}_{t=1}^{N_\text{obs}} (I_t \in \mathbb{R}^{3\times h\times w})$ and depth images $\mathcal{D}=\{D_t\}_{t=1}^{N_\text{obs}} (D_t \in \mathbb{R}^{1\times h\times w})$, TIL aims to localize the hand-object contact timestamps $\mathcal{T}_\text{c}=\{T_i^\text{c}\}_{i=1}^{N_\text{c}}(T_i^\text{c} \in [1,N_\text{obs}])$ and separation timestamps $\mathcal{T}_\text{s}=\{T_j^\text{s}\}_{j=1}^{N_\text{s}}(T_j^\text{s} \in [1,N_\text{obs}])$, which denote the multiple timestamps at which a human hand makes contact with and separates from a target object, respectively. We collectively refer to $\mathcal{T}_\text{c}$ and $\mathcal{T}_\text{s}$ as \textit{interaction transition timestamps}. In this paper, we use the terms \textit{timestamp} and \textit{moment} interchangeably to represent a specific point in time within an input video.
As can be noted, TIL has the potential to trim out the segments of HOI from long egocentric videos precisely, which will facilitate downstream applications such as human-like robot action planning.
Note that we only attend to videos in which only one hand is visible. Besides, due to motion blur issues, rapid non-impulsive actions are not considered in this work. We will explore bimanual HOI processes and rapid actions such as tapping in our future work.

\subsection{Overall Pipeline of EgoLoc}
\label{sec:so}

The overall TIL pipeline of EgoLoc is illustrated in Fig.~\ref{fig:pipeline}, where we take one successful localization of a hand-object contact timestamp as an example. In the self-adaptive sampling strategy (see Fig.~\ref{fig:pipeline}(a)), we first select the anchor frame of the interaction transition timestamps. Anchor frames serve as guidance for subsequent localization of more accurate contact/separation timestamps. The hand motion extraction module extracts 3D hand dynamics from input egocentric observations. This facilitates the following weighted sampling to determine the anchor frame along with its boundary frames. Afterwards, the VLM-based localization module (see Fig.~\ref{fig:pipeline}(b)) introduces the VLM discriminator to identify the interaction attribute, i.e., whether the sampled anchor frame corresponds to a contact timestamp or a separation timestamp. If neither condition holds, we activate the resampling operation in Fig.~\ref{fig:pipeline}(a). Besides, we tile the neighbors of the anchor frame to construct a grid image. The VLM localizer takes as input the grid image to generate the first-round TIL result. Subsequently, the VLM checker in the closed-loop feedback mechanism (see Fig.~\ref{fig:pipeline}(c)) assesses the rationality of the first-round result through visual cues, triggering the second-round inference to refine TIL output once the first-round result is rejected. The VLM discriminator, VLM localizer, and VLM checker share the same out-of-the-box VLM, such as GPT-4o. By iterating the above steps, we can ultimately localize all the possible interaction transition timestamps in any long untrimmed video.

\subsection{Self-Adaptive Sampling Strategy}
\label{sec:sas}

It is non-trivial to let VLM distinguish HOI states with frame-by-frame input of the entire long video. Therefore, we need to conduct high-quality and compact visual prompts, in which the sampled images delineate the approximate localization scope affecting the rationality of the final results. The prior TAL approach~\cite{wake2024open} uniformly samples images from the entire video, leading to unstable localizations and low inference efficiency for long untrimmed videos. In contrast, in this work, we propose a self-adaptive sampling strategy based on hand dynamics analysis, to extract the anchor frame. It facilitates the following interaction attribute identification and high-quality visual prompt construction for VLM-based TIL reasoning. 

\begin{figure*}
  \centering
  \captionsetup{aboveskip=2pt, belowskip=0pt}
  \includegraphics[width=1\linewidth]{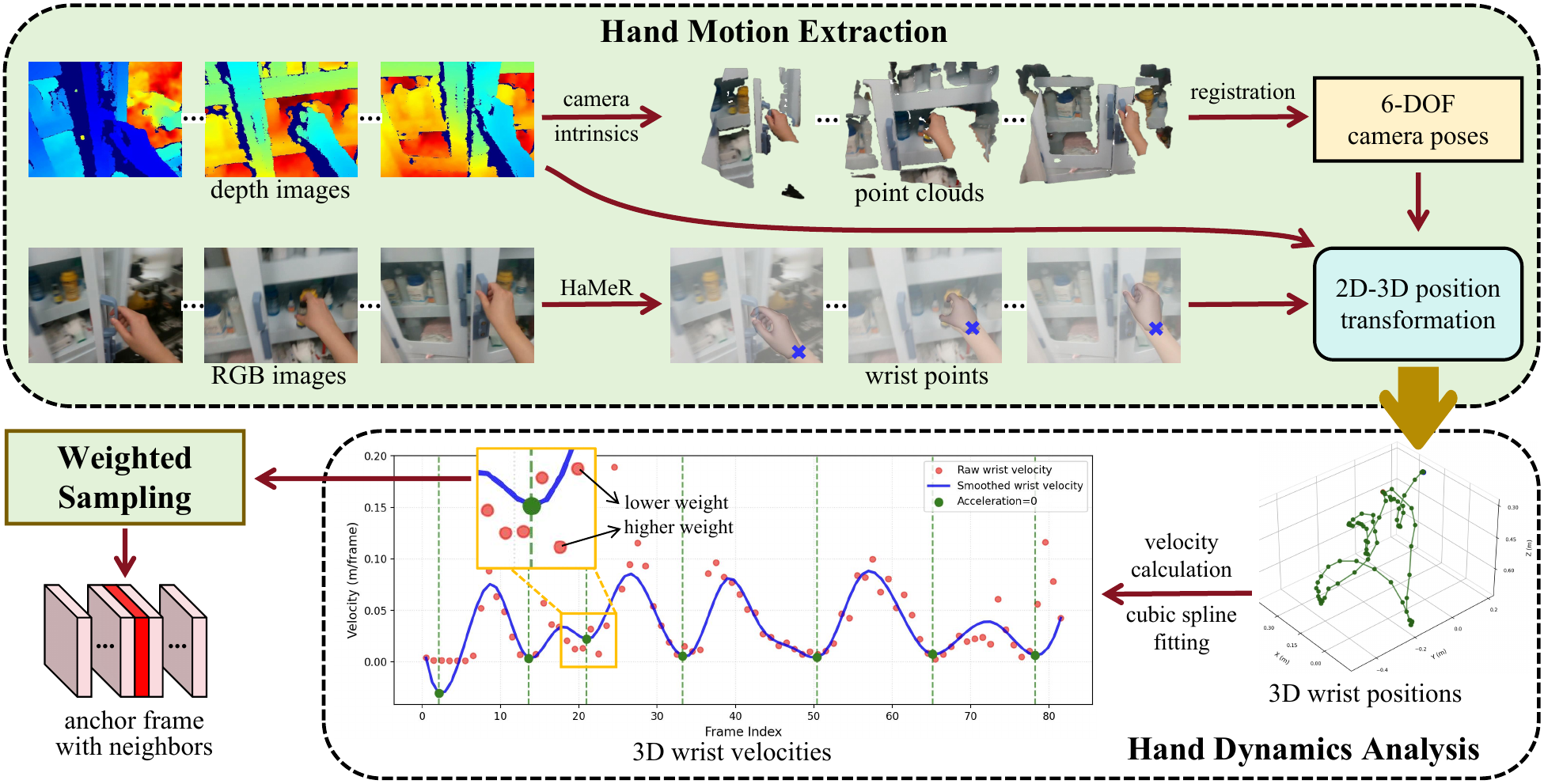}
  \caption{Hand motion extraction and hand dynamics analysis in our devised self-adaptive sampling strategy. We accommodate both 2D and 3D observations to extract 3D hand wrist positions, and calculate wrist velocities and accelerations to produce anchor candidates for the following weighted sampling operation.}
  \label{fig:hand_motion_extraction}
  \vspace{-0.3cm}
\end{figure*}

\subsubsection{{Hand Motion Extraction}}
\label{sec:hme}
We first capture 3D hand dynamics through the hand motion extraction module, which accommodates both 2D and 3D egocentric observations to obtain precise 3D velocities and accelerations of the hand wrist. We attend to the wrist because it is easily detected and exhibits stable motion, whereas other joints are frequently occluded and fail to reflect the global hand motion trends. As Fig.~\ref{fig:hand_motion_extraction} shows, we exploit HaMeR~\cite{pavlakos2024reconstructing} to extract 2D positions of the wrist keypoints $\mathcal{H}^{\text{2D}}=\{H^{\text{2D}}_t\}_{t=1}^{N_\text{obs}} (H^{\text{2D}}_t \in \mathbb{R}^{2})$ on the sequential images $\mathcal{I}$. Then, with the camera intrinsics and the sequential depth images $\mathcal{D}$, we calculate 3D wrist positions $\mathcal{H}^{\text{3D,cam}}=\{H_t^{\text{3D,cam}}\}_{t=1}^{N_\text{obs}} (H_t^{\text{3D,cam}} \in \mathbb{R}^{3})$ in the camera coordinate systems of all timestamps. Next, we perform registration between sequential 3D point clouds from $\mathcal{D}$, obtaining the transformation matrix $\mathcal{M}=\{M_t\}_{t=1}^{N_\text{obs}} (M_t \in \mathbb{R}^{4 \times 4})$. $M_t$ denotes the relative 6-DOF camera pose between the $t$\,th frame and the first frame $(t=1)$. $\mathcal{M}$ transform $\mathcal{H}^{\text{3D,cam}}$ to $\mathcal{H}^{\text{3D,glob}}=\{H_t^{\text{3D,glob}}\}_{t=1}^N (H_t^{\text{3D,glob}} \in \mathbb{R}^{3})$ in the camera coordinate system of the first frame, which is regarded as the global coordinate system in this work. It is worth noting that 3D hand motion is represented in a fixed global frame, eliminating the ambiguity caused by camera egomotion.

Afterwards, as the hand dynamics analysis showcased in Fig.~\ref{fig:hand_motion_extraction}, we use $\mathcal{H}^{\text{3D,glob}}$ to calculate the 3D hand velocities $\mathcal{V} = \{V_t\}_{t=1}^{N_\text{obs}} (V_t \in \mathbb{R}^{1})$ for each timestamp of the egocentric video by $V_t=(H_{t+1}^{\text{3D,glob}}-H_{t}^{\text{3D,glob}})/\delta$, where $\delta$ is the fixed time interval between adjacent frames. We simply set $V_N=V_{N-1}$ since there is no $H^{\text{3D,glob}}_{N+1}$. Then we apply the Savitzky–Golay filter~\cite{savitzky1964smoothing} to $\mathcal{V}$ and fit the velocities with a cubic spline. In this work, we build upon the observation that a hand typically has a relatively lower velocity with an acceleration close to zero when it makes contact with or separates from the target object~\cite{lukos2007choice,cieslik2022research}.
Thus, we extract the local minima of the velocity cubic spline, which have zero hand accelerations. EgoLoc will iteratively loop through all zero-acceleration points to extract all possible interaction transition timestamps. 

\begin{figure*}
  \centering
  \captionsetup{aboveskip=2pt, belowskip=0pt}
  \includegraphics[width=1\linewidth]{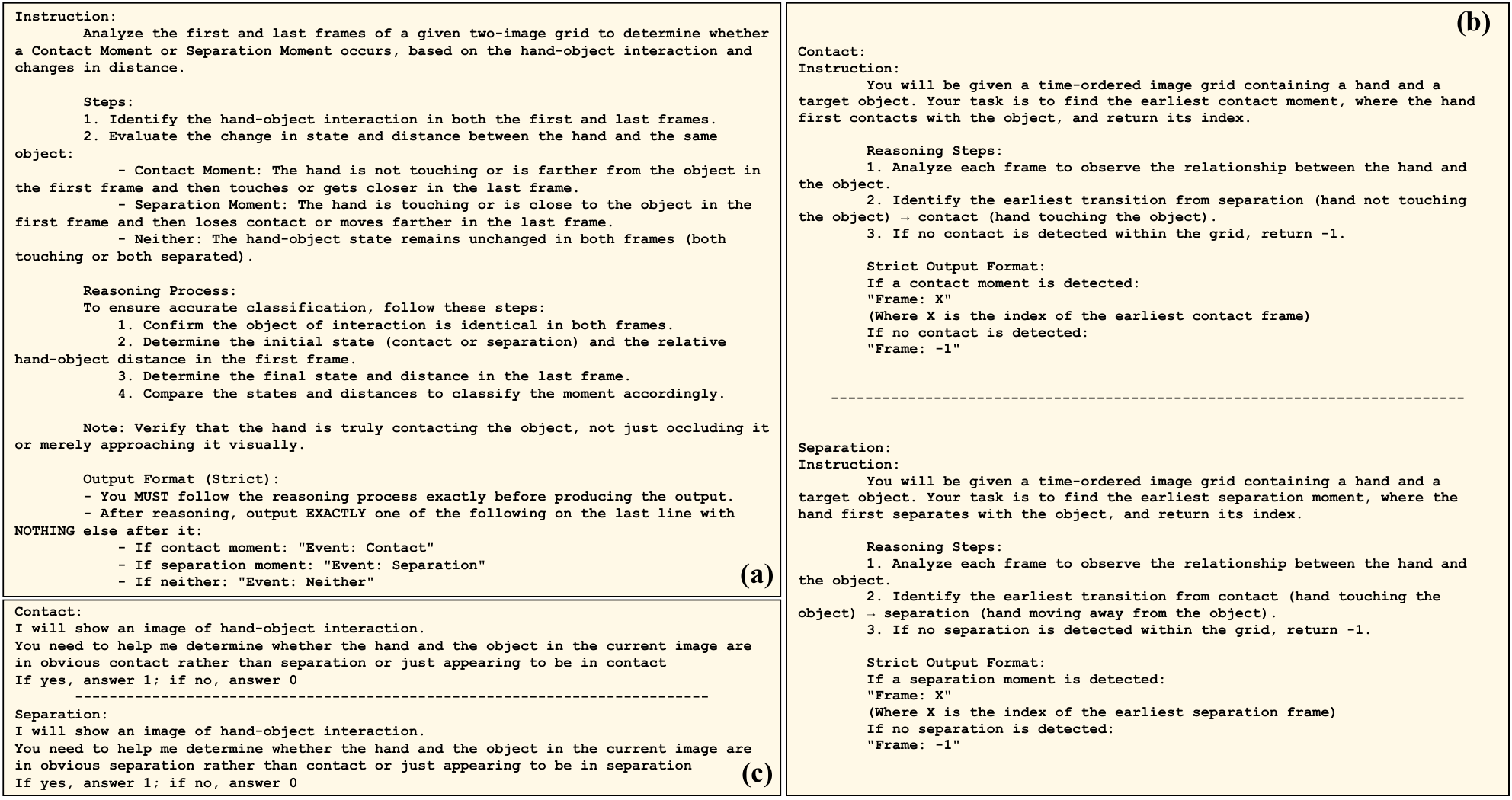}
  \caption{Text prompts tailored for the VLM discriminator (a), VLM localizer (b), and VLM checker (c) in our EgoLoc paradigm. We introduce chain-of-thought (CoT) prompts~\cite{wei2022chain} to improve reasoning accuracy. Notably, these text prompts are universal with respect to different HOI scenes, showing better generalizability compared to the existing TAL/mask-based paradigms limited by predefined action/object categories.}
  \label{fig:text_prompt}
  \vspace{-0.5cm}
\end{figure*}

\subsubsection{{Weighted Sampling}}
\label{sec:ws}

The timestamps of zero accelerations do not strictly align with the frames' timestamps, as showcased in the orange bounding box on the 3D wrist velocities of Fig.~\ref{fig:hand_motion_extraction}. This is because the continuous velocity curve is fitted from separate velocity points. 
Therefore, we harness these zero-acceleration timestamps as ``lighthouses'' to find specific interaction transition timestamps. 
To this end, for each local velocity minimum, we first extract the set of \textit{anchor candidates} $\mathcal{I}^\text{ac}$, which includes the symmetrical $N_\text{ac}$ frames around the zero-acceleration timestamp (included in the orange bounding box in Fig.~\ref{fig:hand_motion_extraction}). The timestamps of these anchor candidates indicate high probabilities of hand-object contact and separation occurring in their vicinity.
Afterwards, we adaptively sample one anchor frame $I_k^\text{ac}$ out of the candidates $\mathcal{I}^\text{ac}$. In the subsequent VLM-based localization module, we will identify the interaction attribute around the sampled anchor frame and construct a grid image with its neighbors as the high-quality visual prompt for the first-round TIL. The sampling weight for the candidate in $\mathcal{I}^\text{ac}$ with the velocity $V_m$ is calculated by:
\begin{align}
    \omega(V_m) = \frac{e^{-\lambda V_m}}{\sum_{n=1}^{N_{\text{ac}}} e^{-\lambda V_n}}
\end{align}
where $\lambda$ is a constant to control the sensitivity of the weights to hand velocities. As a result, lower hand velocities correspond to higher sampling weights, following the observation of hand motion habits mentioned in Sec.~\ref{sec:hme}. After sampling $I_k^\text{ac}$, it is excluded from $\mathcal{I}^\text{ac}$ to prevent repeat sampling in the possible second-round localization. We will switch to the next timestamp with a zero acceleration once one valid interaction transition timestamp is determined within the current $\mathcal{I}^\text{ac}$. As can be seen, we prioritize the higher-likelihood candidates in our self-adaptive sampling strategy to enhance the overall EgoLoc inference efficiency. 
Note that we do not directly regard the anchor frames as localization results, as hand dynamics are not perfectly synchronized with the HOI process.
Besides, their interaction attributes (contact or separation) are hard to recognize with only one image. It is necessary to conduct the visual prompt with more image frames for the following VLM-based localization module to capture the fine-grained HOI state transition. 
Some rare events do not have a local velocity minimum due to the original hand velocities being strictly monotonic. In these cases, the anchor candidates are uniformly sampled from the input video with an interval that is the same as the number of frames in the following grid image.
It is worth noting that we do not analyze 2D hand dynamics in the image plane, considering depth ambiguity and scale aliasing from 2D egocentric observations.

\subsection{VLM-Based Localization Module}
\label{sec:vbr}

The VLM-based localization module aims to localize the precise contact/separation timestamps around the anchor frame from the self-adaptive sampling strategy. It is developed based on off-the-shelf VLMs, encompassing a VLM discriminator and a VLM localizer. They incorporate the visual and text prompts to identify the interaction attribute and output the first-round TIL result, respectively.

\begin{figure}
  \centering
  \captionsetup{aboveskip=2pt, belowskip=0pt}
  \includegraphics[width=0.9\linewidth]{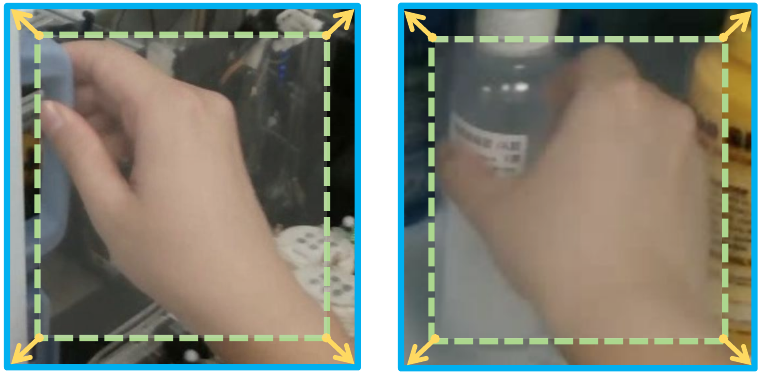}
  \caption{Illustration of hand detection by GroundingDINO. We expand the original bounding boxes for richer target object observations. The green dashed line denotes the original bounding box, while the blue solid line represents the expanded region.}
  \label{fig:bbox_extend}
  \vspace{-0.5cm}
\end{figure}

\subsubsection{{Interaction Attribute Identification}}

Although our self-adaptive sampling strategy produces the anchor frame of high quality compared to the uniform sampling counterpart~\cite{wake2024open}, it is non-trivial to identify whether the anchor frame corresponds to a contact or separation timestamp using only one image as the visual prompt. Therefore, we first propose a VLM discriminator to capture the fine-grained hand-object state transition for interaction attribute identification.

Concretely, we sample $N_\text{adj}^2-1$ adjacent frames associated with the symmetry of $I_k^\text{ac}$ to obtain the enriched initial guesses $\mathcal{I}_k^\text{adj}$. We attend to $\mathcal{I}_k^\text{adj}$ for the following VLM reasoning since it is highly likely that the optimal interaction transition timestamp is located at or near $I_k^\text{ac}$ as mentioned before. Subsequently, we implement GroundingDINO~\cite{liu2023grounding} on the boundary frames, i.e., the temporally first and the last frame in $\mathcal{I}_k^\text{adj}$, given the prompt \textit{hand}. This leads to the region of hand bounding boxes $B_{k,1}^\text{adj}$ and $B_{k,N_\text{adj}^2}^\text{adj}$ (see Fig.~\ref{fig:pipeline}(b), bottom left). As can be noted, the extracted regions in the boundary frames showcase more detailed HOI state transitions, mitigating the distraction from the cluttered background. Only using the boundary frames rather than the entire image sequence $I_k^\text{ac}$ also highlights the changes in HOI states, while reducing disturbance from consecutive visually similar frames. Then we expand both $B_{k,1}^\text{adj}$ and $B_{k,N_\text{adj}^2}^\text{adj}$ by $\epsilon_w$ and $\epsilon_h$ pixels in width and height (see Fig.~\ref{fig:bbox_extend}) to enrich the visual information of the target object. Afterwards, the VLM discriminator takes as input the expanded hand regions, temporally concatenated as the visual prompt, as well as the preset text prompt (see Fig.~\ref{fig:text_prompt}(a)), to identify whether the initial guesses $\mathcal{I}_k^\text{adj}$ correspond to the \underline{contact} timestamp, \underline{separation} timestamp, or \underline{neither}. \underline{neither} denotes that there is no obvious hand-object interaction transition from $B_{k,1}^\text{adj}$ to $B_{k,N_\text{adj}^2}^\text{adj}$, which leads to resampling one anchor frame $I_k^\text{ac}$ from the candidates $\mathcal{I}^\text{ac}$ in Fig.~\ref{fig:pipeline}(a). Otherwise, we get the interaction attribute \underline{contact}/\underline{separation} to produce the text prompt for the subsequent VLM localizer. As detailed and accurate prompts are the foundational basis for VLM reasoning, we argue that explicitly determining interaction attributes to generate attribute-specific text prompts enhances the reasoning coherence of the VLM localizer.

\subsubsection{{First-Round TIL}}

Inspired by~\cite{wake2024open}, we then construct a grid image $G_k \in \mathbb{R}^{N_\text{adj}h\times N_\text{adj}w}$ as the visual prompt for the VLM localizer.
Specifically, we tile all the $N_\text{adj}^2$ frames encompassing $I_k^\text{ac}$ and its neighbors in $\mathcal{I}_k^\text{adj}$, which are automatically annotated in chronological order. The right side of Fig.~\ref{fig:pipeline}(b) illustrates an example of the grid image $G_k$ with $N_\text{adj}=4$, where $I_k^\text{ac}$ holds the index number 8. Subsequently, we feed the grid image and the attribute-specific text prompt (see Fig.~\ref{fig:text_prompt}(b)) into the VLM localizer. It localizes the contact/separation moment $\hat{T}_i^\text{c}$ or $\hat{T}_j^\text{s}$ within the tiled $\mathcal{I}_k^\text{adj}$ as the first-round TIL result. Notably, our VLM localizer performs single-step inference rather than multiple inference iterations. This is because the visual prompt $G_k$ is elaborated by our self-adaptive sampling strategy, which offers high-quality initial guesses compared to the coarsely sampled counterparts~\cite{wake2024open}. This significantly improves the temporal localization efficiency.

\subsection{Closed-Loop Feedback Mechanism}
\label{sec:clf}

The VLM discriminator implements a simpler task than the VLM localizer since interaction attribute identification with two boundary frames is much more tractable than temporal localization within multiple tiled images. Therefore, the VLM localizer is more prone to hallucinations, leading to incorrect first-round results. To address this issue, we further propose a closed-loop feedback mechanism to refine the contact timestamp $\hat{T}_i^\text{c}$ and separation timestamp $\hat{T}_j^\text{s}$ estimated by the VLM localizer. Through our devised closed-loop feedback, EgoLoc achieves self-improvement by incorporating visual cues with in-context learning, resulting in more accurate TIL results than the open-loop schemes.

\subsubsection{{Visual Cue Assessment}}

In our paradigm, the closed-loop feedback specifically refers to the justification of whether VLM-based localization should be re-implemented. It is generated by assessing the visual cues of the first-round contact/separation moments. We directly feed the egocentric image of the first-round TIL result along with the text prompt (see Fig.~\ref{fig:text_prompt}(c)) to the VLM checker. It determines whether there are obvious visual mistakes in the first-round result, and outputs whether the first-round result can be accepted as the final TIL result (\underline{Yes}/\underline{No}) of the current iteration. For instance, when the interaction attribute identified by the VLM discriminator is ``contact'' but the first-round result from the VLM localizer visually exhibits a clear hand-object separation, the VLM checker can identify this discrepancy and outputs \underline{No}.

\begin{figure}
  \centering
  \captionsetup{aboveskip=2pt, belowskip=0pt}
  \includegraphics[width=1\linewidth]{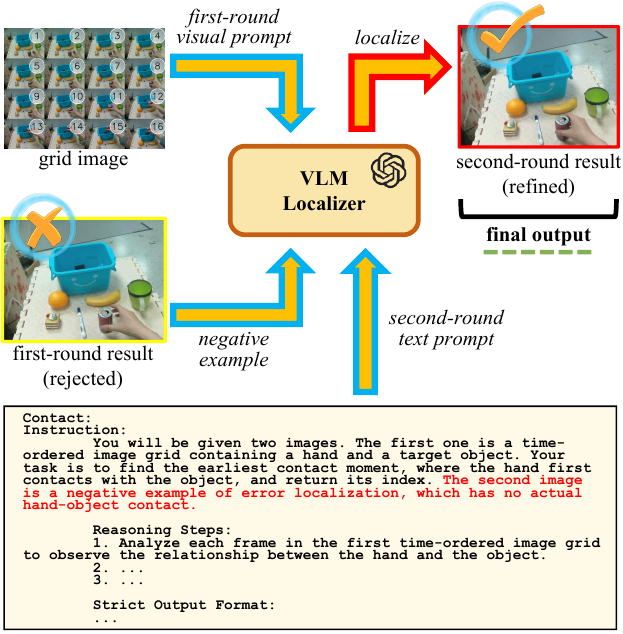}
  \caption{Illustration of in-context learning leveraged in our closed-loop feedback mechanism (taking the refinement of contact timestamps as an example). The rejected first-round result is regarded as a negative in-context example to mitigate VLM hallucination. The omitted parts in the second-round text prompt are the same as the first-round counterparts for the VLM localizer.}
  \label{fig:in_context}
  \vspace{-0.5cm}
\end{figure}

\subsubsection{{Second-Round TIL}}

The answer \underline{No} from the visual cue assessment triggers implementing in-context learning~\cite{brown2020language} for re-localization by the VLM localizer in the VLM-based localization module. As illustrated in Fig.~\ref{fig:in_context}, we add the wrong first-round image frame as an additional negative example in the prompt for the VLM localizer, to mitigate its hallucination. As a result, the VLM localizer receives the enriched text prompt, the first-round grid image, and the negative in-context example to implement the second-round TIL reasoning. We therefore close the TIL loop and obtain refined contact and separation moment estimation $\hat{T}_i^\text{c*}$ and $\hat{T}_j^\text{s*}$ as the second-round TIL result. EgoLoc does not rely on a rule-based feedback design with limited generalizability, but instead flexibly utilizes the first-round results as guidance for the second-round refinement.
The ablation studies in Sec.~\ref{sec:exp_albation} will demonstrate that our devised closed-loop feedback mechanism effectively reduces estimation uncertainties and improves localization accuracy. 
We empirically observed that one round of closed-loop feedback notably improves TIL performance, whereas additional rounds offer limited benefits and reduce inference efficiency. Thus, EgoLoc exploits a single feedback round for a practical trade-off.

As can be noted, our proposed paradigm effectively resolves the inefficiency inherent in frame-by-frame and uniform sampling approaches, while maintaining the computational advantages of adaptive sampling-based localization.
It also incorporates a closed-loop feedback mechanism that substantially reduces inference uncertainties arising from VLM hallucinations in tiled multiple frames, thereby producing more accurate and refined TIL outputs.
To find all the interaction transition timestamps in any long untrimmed video, EgoLoc iteratively loops through the zero-acceleration timestamps, implements weighted sampling in the self-adaptive sampling strategy, performs first-round localization in the VLM-based localization module, and achieves second-round refinement with the closed-loop feedback mechanism.
EgoLoc is a hand-centric reasoning framework that is not affected by inaccurate mask generation for target object instances. Its holistic inference process can also be automatically performed without pertaining with manual annotations. Thus, we achieve generalizable temporal interaction localization in a zero-shot manner for any egocentric videos.

\begin{figure}
  \centering
  \captionsetup{aboveskip=2pt, belowskip=0pt}
  \includegraphics[width=1\linewidth]{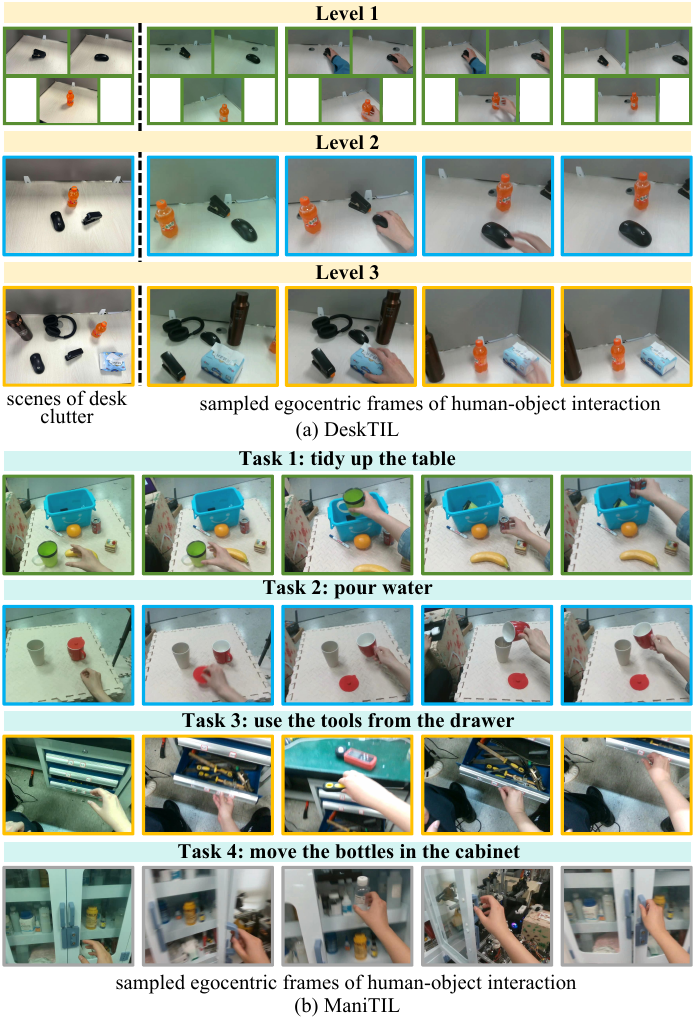}
  \caption{Visualization of the egocentric HOI data in our newly proposed DeskTIL and ManiTIL benchmarks.}
  \label{fig:benchmark}
  \vspace{-0.5cm}
\end{figure}

\section{Experiments}
\label{sec:experiments}

\begin{figure*}
  \centering
  \captionsetup{aboveskip=2pt, belowskip=0pt}
  \includegraphics[width=1\linewidth]{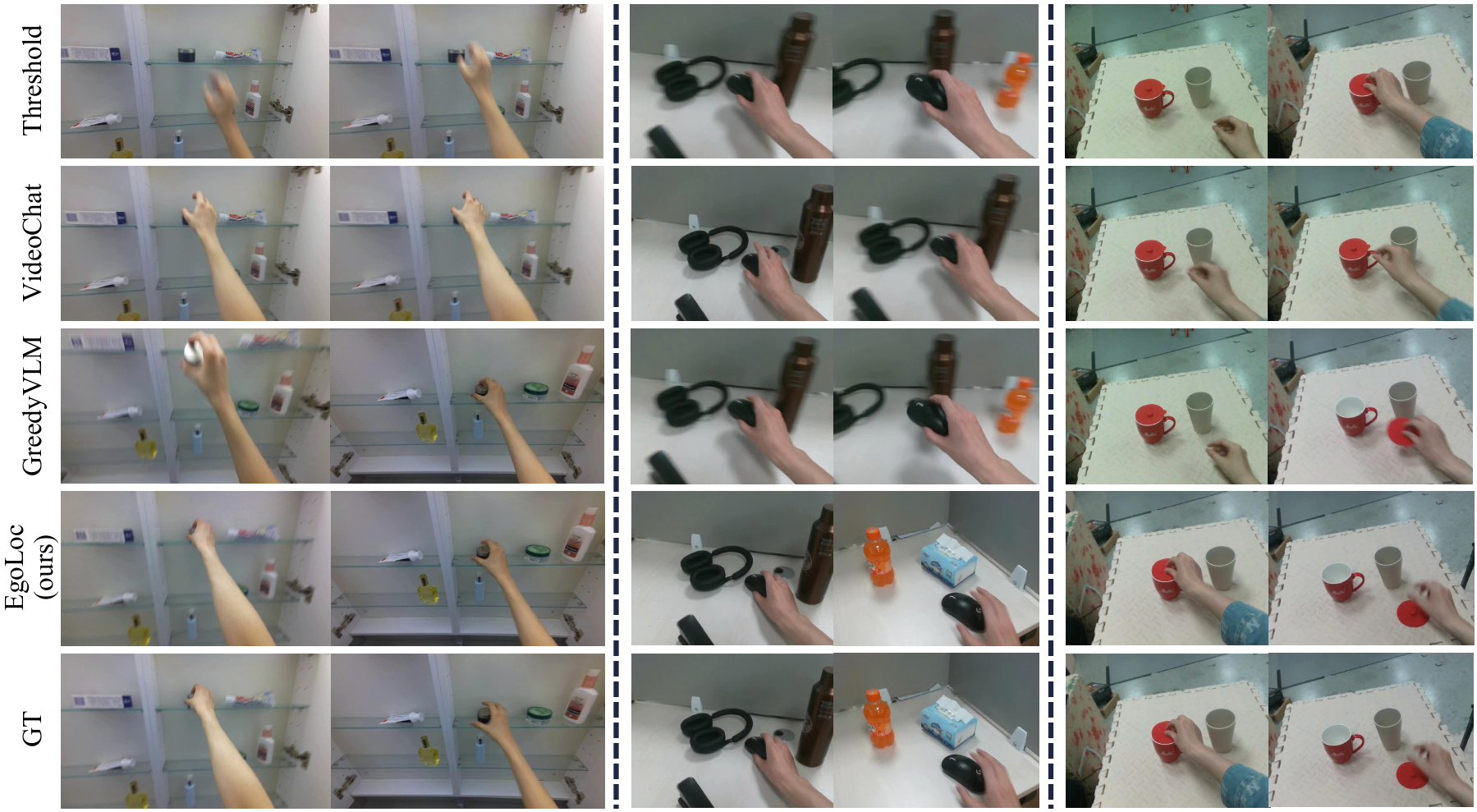}
  \caption{Visualization of TIL results from EgoLoc, baselines, and GT annotations (left: EgoPAT3D-DT, middle: DeskTIL, right: ManiTIL).}
  \label{fig:main_results}
  \vspace{-0.5cm}
\end{figure*}

\subsection{Implementation Details}

\subsubsection{{Benchmarks and Baselines}}

We first utilize the EgoPAT3D dataset~\cite{li2022egocentric} to evaluate the performance of temporal interaction localization. In particular, we use the EgoPAT3D-DT subset processed by Bao~\etal~\cite{bao2023uncertainty}, which contains finer-grained hand-object interaction sequences. 
We adopt their original frame rate 30\,FPS, and manually annotate the contact and separation timestamps for each video.
Moreover, we propose two novel TIL benchmarks, \textbf{DeskTIL} and \textbf{ManiTIL}, for more comprehensive assessment. DeskTIL attends to different levels of desk clutter (see Fig.~\ref{fig:benchmark}(a)). We use a head-mounted RealSense D435i to collect 50 video sequences per level, with an average duration of around 6\,s per sequence. Each video in DeskTIL contains a single HOI process (one pre-contact stage, one in-contact stage, and one post-contact stage), and is downsampled to 5\,FPS for TIL evaluation under a low frame rate constraint. To further validate the TIL performance on longer videos, we build ManiTIL that encompasses four long-term HOI tasks: \textit{tidy up the table}, \textit{pour water}, \textit{use the tools from the drawer}, and \textit{move the bottles in the cabinet} (see Fig.~\ref{fig:benchmark}(b)). We collect 50 videos for each task at 30\,FPS. Each video in ManiTIL includes more than one HOI process, which indicates that the TIL methods are required to localize multiple contact timestamps and separation timestamps. Besides, each task has different numbers of interaction transitions. For instance, in the task \textit{tidy up the table}, we record storing objects in varying ratios on the table. We also manually label the interaction transition timestamps for both DeskTIL and ManiTIL benchmarks. Supp. Mat., Sec.~A introduces their statistical properties regarding hand motion distributions. We will release them as open source to facilitate future work in the literature.

We select {T-PIVOT}~\cite{wake2024open} as one of our primary baselines, as it is the pioneering zero-shot method achieving SOTA ZS-TAL performance. We adapt it to TIL by providing the same text prompt as the one for our VLM localizer. Besides, a powerful chat-centric video understanding approach, {VideoChat (LongVideo version)}~\cite{li2024mvbench,li2023videochat} is also utilized as a VLM-based baseline. We provide devised text prompts to let it retrieve interaction transition timestamps within input videos.
We also design a new TIL baseline, namely {GreedyVLM}. Without any sampling operation, it directly uses the holistic image sequence to construct the grid image for each video, which is then fed to the VLM to output contact and separation timestamps. 
In addition to these VLM-based methods, we conduct an ad-hoc baseline {Threshold} referring to the previous works~\cite{haldar2025point,liu2025egozero}, which simply thresholds the Euclidean distance between the tips of the index finger and thumb to detect interaction transition timestamps.
Moreover, we introduce a mask-based baseline {HOIMask}~\cite{ren2025motion}, which localizes hand-object contact and separation according to the number of thumb keypoints plus any one of the other fingertips within the target object mask. We report its performance on our ManiTIL since only this benchmark provides specific task descriptions for target object mask generation.

\subsubsection{{Evaluation Metrics}}
\label{sec:eval_metrics}
We exploit the metrics MoF and IoU widely used in the TAL literature, to evaluate the performance of temporal interaction localization in this work. Moreover, considering the fine-grained nature of the TIL task, we introduce two new metrics, MAE and SR, to comprehensively quantify the accuracy of contact and separation timestamp estimations. These metrics are detailed as follows.
\begin{itemize}[leftmargin=1em]
\item \textbf{MoF (Mean over Frames)} represents the percentage of the frames within correctly estimated hand-object in-contact stages out of the total in-contact frames for each video. We will report the average MoF across all videos.

\item  \textbf{IoU (Intersection over Union)} measures the overlap between the estimated and ground-truth segments of the hand-object in-contact stage for each video. It is calculated as the ratio of the intersection to the union of the two frame sets. We will report the average value of this metric across all videos.

\item  \textbf{MAE (Mean Absolute Error)} measures the absolute error, i.e., the frame interval between the estimated and ground-truth interaction transition timestamps for each video. We will report the average value of this metric across all videos.

\item  \textbf{SR (Success Rate)} calculates the proportion of successfully matched contact/separation timestamps between the estimations and ground-truth annotations considering all video clips. We set different tolerance ranges $\gamma$ for this metric. One estimation can be regarded as a success if the frame interval between the estimated and ground-truth timestamps falls within the preset tolerance range.

\end{itemize}

\subsubsection{{EgoLoc Details}}
\label{sec:config}

In the self-adaptive sampling strategy, the frame
time interval $\delta$ is $1/30$\,s for EgoPAT3D-DT and ManiTIL, and $1/5$\,s for DeskTIL. The number of anchor candidates $N_\text{ac}$ is set as $5$. In the VLM-based localization module, $\epsilon_w=\epsilon_h=10$ is utilized to expand hand regions for interaction attribute identification. Besides, we set $N_\text{adj}$ to $2\sim4$ to construct the grid image in the following experiments.
We let the VLM discriminator, VLM localizer, and VLM checker share the same GPT-4o by default. 
For EgoLoc and all the other baselines, we perform 5 TIL trials for each video and report the averaged evaluation metrics across them.

\begin{table*}[t]
\setlength{\tabcolsep}{15pt}
\center
\renewcommand\arraystretch{0.9}
\caption{Comparison of performance on temporal interaction localization on EgoPAT3D-DT, DeskTIL, and ManiTIL. Best and secondary results are viewed in \textbf{bold black} and \myblue{blue} colors respectively.}
\vspace{-0.15cm}
\begin{tabular}{l|cccccc}
\toprule
\multicolumn{1}{l|}{\multirow{2}{*}{Approach}}   & \multicolumn{6}{c}{EgoPAT3D-DT}  \\ \cmidrule{2-7} 
\multicolumn{1}{c|}{}                                                                               & SR($\gamma=1$)\,$\uparrow$    & SR($\gamma=3$)\,$\uparrow$ & SR($\gamma=5$)\,$\uparrow$    & MAE\,$\downarrow$ & MoF\,$\uparrow$   & IoU\,$\uparrow$   \\ \cmidrule{1-7}     
Threshold    & 0.137     &0.255	 & 0.382    & 12.559	  &0.312    & 0.128 \\
VideoChat  & 0.105     &0.211	 & 0.368    & 8.408	  &0.586    & 0.217 \\
GreedyVLM    &0.137      &0.324      &0.529          &5.794    &0.655    &0.565\\
T-PIVOT ($N_\text{adj}=2$)   & 0.225    &0.255 	 &0.451    &11.284 	  &0.476   &0.324  \\ 
T-PIVOT ($N_\text{adj}=3$)   & 0.216     & 0.402	 & 0.647    & 8.176	  &0.580   & 0.469 \\ 
T-PIVOT ($N_\text{adj}=4$)   & 0.294     &0.500	 & \myblue{0.755}   & 6.471	  &0.650    & 0.550 \\  
Ours ($N_\text{adj}=2$) & \textbf{0.362}   & \textbf{0.628} 	& \textbf{0.819}  	& \textbf{3.713}	 & \textbf{0.784} 	& \textbf{0.726}\\ 
Ours ($N_\text{adj}=3$) & \myblue{0.298}    & \myblue{0.595} 	& {0.726}  	& \myblue{4.417}	 & {0.714} 	& {0.645}\\
Ours ($N_\text{adj}=4$) & {0.272}    & {0.568} 	& {0.705}  	& {4.568}	 & \myblue{0.729} 	& \myblue{0.657}\\
\midrule
\multicolumn{1}{l|}{Approach}   & \multicolumn{6}{c}{DeskTIL}  \\ \midrule
Threshold    & 0.333     &0.482	 & 0.546    & 5.407	  &0.707    & 0.302 \\
VideoChat  & 0.407     &0.602	 & 0.704    & 3.935	  &0.747    & 0.441 \\
GreedyVLM      &0.343       &0.611       &0.815        &3.361    &0.773    &0.556\\
T-PIVOT ($N_\text{adj}=2$)   &  \myblue{0.450}  & \myblue{0.833} 	 &0.867    & 2.333	    & 0.835 &  \myblue{0.715} \\
T-PIVOT ($N_\text{adj}=3$)   & 0.417     & 0.750	 & 0.833     & 3.000	 & 0.800  & 0.706
   \\ 
T-PIVOT ($N_\text{adj}=4$)   & 0.333     &0.667	 & 0.787     & 3.428	  &0.789    & 0.692 \\ 
Ours ($N_\text{adj}=2$) & \textbf{0.573}   & \textbf{0.927} 	& \textbf{1.000}     	& \textbf{0.611}	 & \textbf{0.940} 	& \textbf{0.849}
     \\ 
Ours ($N_\text{adj}=3$) & \myblue{0.450}    & 0.763 	& \myblue{0.950}  	& \myblue{2.175}	 & \myblue{0.856}	& 0.658
     \\ 
Ours ($N_\text{adj}=4$) & {0.385}    & {0.519} 	& {0.846}    	& {3.154}	 & {0.786} 	& {0.483} \\  \midrule
\multicolumn{1}{l|}{Approach}   & \multicolumn{6}{c}{ManiTIL}  \\ \midrule
Threshold         &0.157	 & 0.339  & 0.507  & 6.367	  &0.097    & 0.139 \\
HOIMask       &0.117	 & 0.286  & 0.465  & 7.089	  &0.220    & 0.286 \\ 
VideoChat       &0.130	 & 0.302  & 0.468  & 7.254	  &0.065    & 0.098 \\
GreedyVLM            &0.135       &0.345     & 0.516    &6.096    &\textbf{0.307}    &\myblue{0.403}\\
Ours ($N_\text{adj}=2$)    & \textbf{0.328} 	& \textbf{0.645}   & \textbf{0.761} 	& \textbf{4.289}	 & \myblue{0.290} 	& \textbf{0.429}
     \\ 
Ours ($N_\text{adj}=3$)    & \myblue{0.276} 	& \myblue{0.551}   & \myblue{0.736} 	& \myblue{4.698}	 & {0.281} 	& {0.399}
     \\ 
Ours ($N_\text{adj}=4$)    & {0.220} 	& {0.466}   & 0.651	& {5.285}	 & {0.259} 	& {0.359}\\
\bottomrule
\end{tabular}
\label{tab:compare_main}
\vspace{-0.2cm}
\end{table*}

\begin{figure}
  \centering
  \captionsetup{aboveskip=2pt, belowskip=0pt}
  \includegraphics[width=1\linewidth]{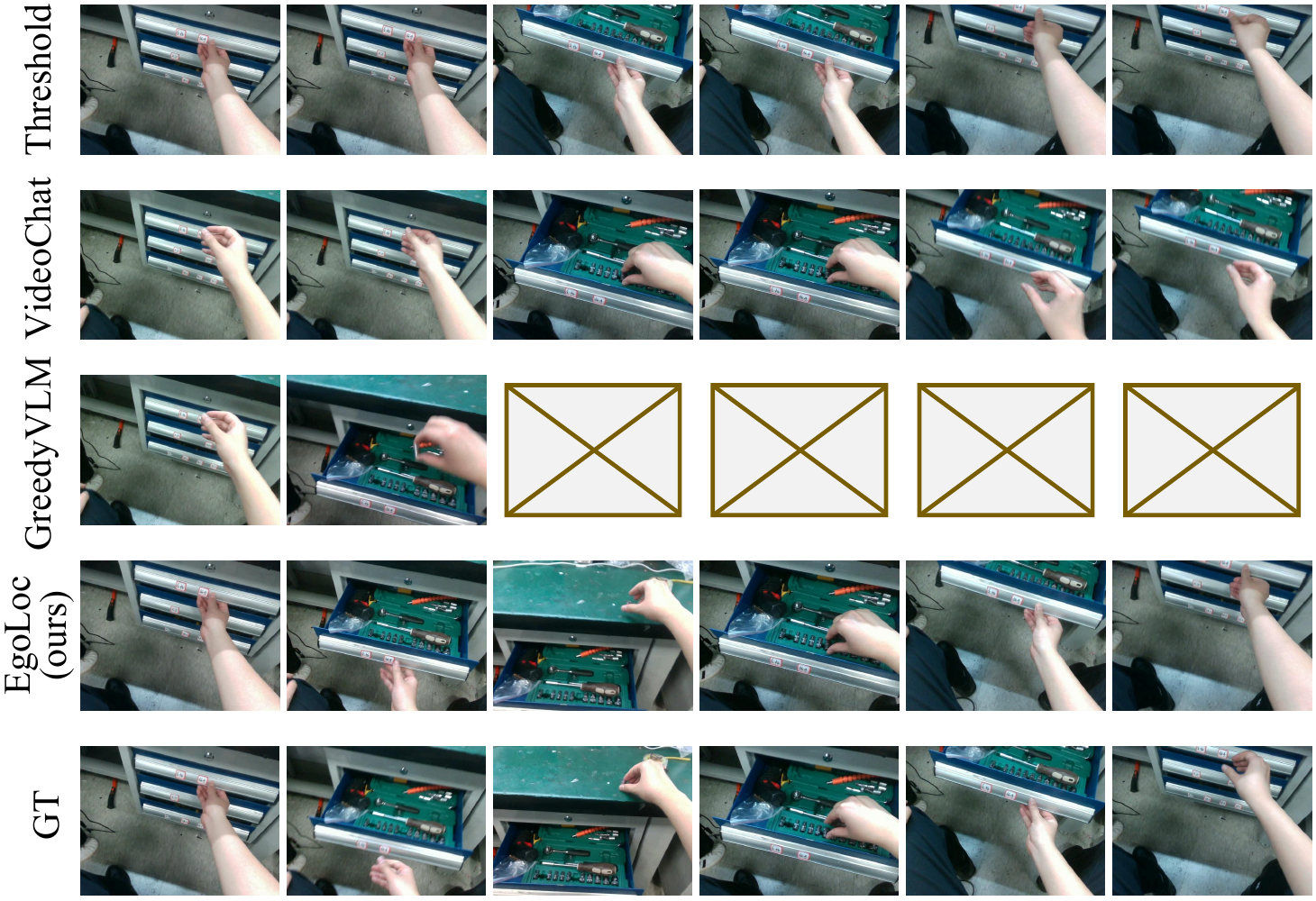}
  \caption{Visualization of TIL results on the example sequence of ManiTIL, which contains multiple HOI processes to complete the task \textit{use the tools from the drawer}. GreedyVLM only identifies two interaction transition timestamps.}
  \label{fig:results_long_video}
  \vspace{-0.5cm}
\end{figure}

\subsection{Main Results}
\label{sec:comparison_with_prior}

\begin{figure}
  \centering
  \captionsetup{aboveskip=2pt, belowskip=0pt}
  \includegraphics[width=1\linewidth]{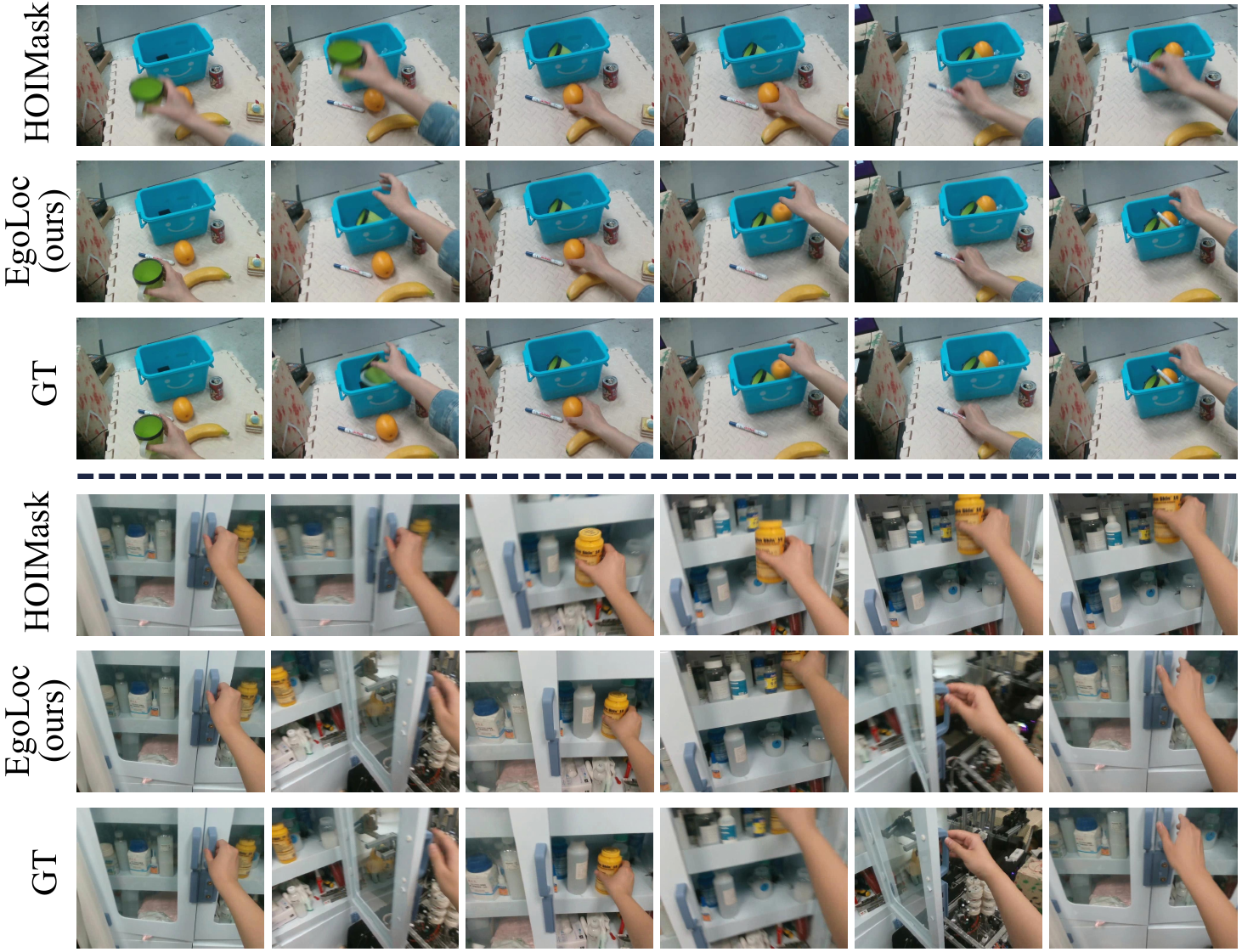}
  \caption{Qualitative comparison between HOIMask and EgoLoc on the example sequence to complete the task \textit{tidy up the table} (upper) and \textit{move the bottles in the cabinet} (lower) in ManiTIL.}
  \label{fig:results_hoi_mask}
  \vspace{-0.5cm}
\end{figure}

We first compare the TIL performance of our proposed EgoLoc method with the selected baselines. 
As shown in Tab.~\ref{tab:compare_main}, our proposed EgoLoc significantly outperforms the baseline methods on nearly all the evaluation metrics for the EgoPAT3D-DT dataset, our DeskTIL benchmark, and ManiTIL benchmark. The baseline Threshold performs poorly across all datasets due to its hard-coded nature. VideoChat also fails to retrieve precise interaction transition timestamps since its retrieval scheme lacks TIL-specific optimization. T-PIVOT adopts a coarse sampling phase for initial guesses, while GreedyVLM attends to brute-force processing of all images at once without any sampling operation. They both present suboptimal performance due to the absence of a reasonable sampling strategy and self-improvement mechanism. In contrast, our proposed EgoLoc exhibits consistently better TIL performance across multiple datasets, thanks to our self-adaptive sampling strategy and closed-loop feedback mechanism. We do not test T-PIVOT on ManiTIL since it is not trivial to adapt it to localize timestamps for multiple HOI stages. It is also noteworthy that EgoLoc with only $N_\text{adj}=2$ adjacent frames of the anchor frame significantly outperforms T-PIVOT with $N_\text{adj}=\{2,3,4\}$. Counterintuitively, a larger $N_\text{adj}$ leads to degraded localization accuracy for EgoLoc. We attribute this to the increasing spatial-temporal receptive field that causes the VLM's attention to distribute weights across irrelevant frames.
In Fig.~\ref{fig:main_results}, we present a qualitative comparison between EgoLoc and the baselines. EgoLoc consistently achieves superior performance in interaction localization, particularly excelling in challenging scenarios with multiple distractors. Fig.~\ref{fig:results_long_video} also showcases that EgoLoc accurately localizes multiple contact/separation timestamps of several HOI processes in long untrimmed videos of ManiTIL. Moreover, Fig.~\ref{fig:results_hoi_mask} visualizes the TIL results of our EgoLoc paradigm and the conventional mask-based scheme on ManiTIL. For the task \textit{tidy up the table}, we use \textit{daily manipulanda} as the text prompt of Grounded SAM~\cite{ren2024grounded} to generate object masks for HOIMask. For the task \textit{move the bottles in the cabinet}, we simultaneously use the prompts \textit{handle} and \textit{bottle} to generate masks for all possible manipulated targets. As can be seen, HOIMask shows poor localization performance, which cannot automatically and accurately select the next active object instances. In contrast, EgoLoc exploits the strong reasoning ability of VLMs, accurately localizing interaction transition timestamps across different HOIs without the requirement of object masks.
More visualizations are provided in Supp. Mat., Sec.~C.
Note that EgoLoc completes a TIL inference in approximately $0.53\times\sim 0.77\times$ the time required by T-PIVOT. This also benefits from our hand-dynamics-guided sampling operation, which endows VLM with a higher-quality localization scope than the uniform sampling in the baseline.

\subsection{Ablation Studies}
\label{sec:exp_albation}
In this section, we ablate the key components proposed in EgoLoc. $N_\text{adj}=2$ is set by default for all the following experiments.

\textbf{Adaptive Sampling}. 
We first evaluate the effectiveness of our proposed self-adaptive sampling strategy (SASS), which incorporates 3D perception data to extract 3D hand dynamics. Specifically, we conduct three baselines in this ablation experiment, including EgoLoc w/o SASS, w/ SASS-2D, and w/ SASS-3D. The baseline w/o SASS replaces EgoLoc's SASS with random sampling, while SASS-2D denotes the baseline only using 2D RGB images as input. SASS-2D implements 2D hand dynamics analysis, with the detected wrist locations in the image plane.
The experimental results in Tab.~\ref{tab:abla_sass} show that incorporating 3D perception information for adaptive sampling significantly improves the baselines on all the evaluation metrics. There is a substantial drop in localization performance when employing random sampling in the baseline without SASS. The localization accuracy also exhibits a degradation when extracting 2D hand dynamics as a sampling reference. These results indicate that our proposed SASS generates high-quality initial guesses with 3D hand dynamic analysis for VLM-based reasoning. Exploiting 3D hand velocities from 3D perception also enhances the model's spatial awareness of actual human motion characteristics, avoiding depth ambiguity and scale aliasing of 2D observations.

\begin{table}[t]
\small
\setlength{\tabcolsep}{7.5pt}
\center
\renewcommand\arraystretch{1}
\caption{Ablation study on the self-adaptive sampling strategy with EgoPAT3D-DT. Best results are viewed in \textbf{bold black}.}
\vspace{-0.1cm}
\begin{tabular}{l|cccc}
\toprule
\multicolumn{1}{l|}{\multirow{1}{*}{Approach}}     & SR($\gamma=5$)\,$\uparrow$    & MAE\,$\downarrow$ & MoF\,$\uparrow$    & IoU\,$\uparrow$    \\ \cmidrule{1-5}  
w/o SASS  	& 0.412  & 16.137  & 0.291   & 0.096 \\
w/ SASS-2D  & 0.726  & 5.288  & 0.682   & 0.594 \\
w/ SASS-3D  & \textbf{0.819}  & \textbf{3.713}   & \textbf{0.784} 	 & \textbf{0.726} \\
\bottomrule
\end{tabular}
\label{tab:abla_sass}
\vspace{-0.3cm}
\end{table}

\begin{table}[t]
\small
\setlength{\tabcolsep}{6pt}
\center
\renewcommand\arraystretch{1}
\caption{Ablation study on the interaction attribute identification with EgoPAT3D-DT. Best results are viewed in \textbf{bold black}.}
\vspace{-0.1cm}
\begin{tabular}{l|cccc}
\toprule
\multicolumn{1}{l|}{\multirow{1}{*}{Approach}}    & SR($\gamma=5$)\,$\uparrow$   & MAE\,$\downarrow$ & MoF\,$\uparrow$    & IoU\,$\uparrow$    \\ \cmidrule{1-5} 
Anchor frame   	& 0.667  & 7.500  & 0.596   & 0.493 \\
Grid image  	& 0.750   & 6.050  & 0.664   & 0.557 \\
Boundary frames & \textbf{0.819}  & \textbf{3.713}   & \textbf{0.784} 	 & \textbf{0.726}	 \\
\bottomrule
\end{tabular}
\label{tab:abla_iai}
\vspace{-0.3cm}
\end{table}

\begin{table}[t]
\small
\setlength{\tabcolsep}{2.6pt}
\center
\caption{Ablation study on the closed-loop feedback mechanism with DeskTIL. Best results are viewed in \textbf{bold black}.}
\vspace{-0.1cm}
\renewcommand\arraystretch{1}
\begin{tabular}{cc|cccc}
\toprule
\makecell[c]{Visual cue \\ assessment} & \makecell[c]{In-context\\learning}  & SR ($\gamma=5$)\,$\uparrow$      & MAE\,$\downarrow$ & MoF\,$\uparrow$    & IoU\,$\uparrow$   \\ \midrule          
\ding{55} &  \ding{55}  & 0.921    & 1.485	 & 0.895 	& 0.755  \\
\ding{51} &\ding{55}     &0.976      &1.402    &0.907  	 & 0.774	 \\ 
\ding{51}  & \ding{51}  	& \textbf{1.000}  & \textbf{0.611}   & \textbf{0.940} 	 & \textbf{0.849}\\ \bottomrule
\end{tabular}
\label{tab:ala_on_feedback1}
\vspace{-0.2cm}
\end{table}

\begin{table}[t]
\small
\setlength{\tabcolsep}{2pt}
\center
\renewcommand\arraystretch{1}
\caption{Ablation study on inference uncertainties with EgoPAT3D-DT. Best results are viewed in \textbf{bold black}.}
\vspace{-0.1cm}
\begin{tabular}{l|cccc}
\toprule
\multicolumn{1}{l|}{\multirow{1}{*}{Approach}}    & SR$^{\dag}$($\gamma=5$)\,$\downarrow$      & MAE$^{\dag}$\,$\downarrow$ & MoF$^{\dag}$\,$\downarrow$    & IoU$^{\dag}$\,$\downarrow$    \\ \cmidrule{1-5}    
VideoChat  	& 0.051  & 0.260  & 0.026   & 0.035\\
GreedyVLM  & 0.026  	& 0.240  & 0.029   & 0.037\\
T-PIVOT  & 0.033	  & 0.922  & 0.033   & 0.045\\
Ours (w/o feedback)    & 0.041	& 0.327  & 0.022   & 0.025 \\
Ours (w/ feedback)  & \textbf{0.018}	  & \textbf{0.138}  & \textbf{0.019}   & \textbf{0.024} \\
\bottomrule
\multicolumn{3}{l}{$\dag$: standard deviation of the metric.} 
\end{tabular}
\label{tab:uncertainty}
\vspace{-0.3cm}
\end{table}

\begin{table}[t]
\small
\setlength{\tabcolsep}{6.2pt}
\center
\renewcommand\arraystretch{1}
\caption{Ablation study on VLMs with EgoPAT3D-DT. Best results are viewed in \textbf{bold black}.}
\vspace{-0.1cm}
\begin{tabular}{l|ccccc}
\toprule
\multicolumn{1}{l|}{\multirow{1}{*}{VLM}}     & SR ($\gamma=5$)\,$\uparrow$    & MAE\,$\downarrow$ & MoF\,$\uparrow$    & IoU\,$\uparrow$    \\ \cmidrule{1-5} 
Janus-Pro-7B  	& 0.469  & 7.500 & 0.592  & 0.482\\
Gemini 2.5 Pro  & 0.798  & 4.586 & 0.727   & 0.654\\
GPT-4o mini  	& 0.678  & 5.633 & 0.676   & 0.588\\
GPT-4o  	& \textbf{0.819}  & \textbf{3.713}   & \textbf{0.784} 	 & \textbf{0.726} \\
\bottomrule
\end{tabular}
\label{tab:aba_vlm}
\vspace{-0.2cm}
\end{table}

\textbf{Interaction Attribute Identification}. We further ablate our proposed interaction attribute identification with boundary frames in $\mathcal{I}_k^\text{adj}$. 
Tab.~\ref{tab:abla_iai} presents the performance comparison when substituting the VLM discriminator's input boundary frames with both the single anchor frame and the entire grid image.
Only using the anchor frame to determine the contact/separation attributes yields the lowest accuracy, as one single static snapshot inherently fails to capture the temporal dynamics of hand-object state transitions. Besides, when exploiting the entire grid image with $N_\text{adj}^2$ frames, the TIL performance also drops compared to using boundary frames $B_{k,1}^\text{adj}$ and $B_{k,N_\text{adj}^2}^\text{adj}$. This suggests that the utilization of two boundary frames addresses the potential hallucination issue in interaction attribute identification with multiple visually similar frames.

\textbf{Closed-Loop Feedback}.
Next, we ablate our proposed closed-loop feedback mechanism of EgoLoc. We present the TIL performance of the baselines including EgoLoc without closed-loop feedback, with only visual cue assessment to trigger second-round localization, and with both visual cue assessment and in-context learning. 
As shown in Tab.~\ref{tab:ala_on_feedback1}, the combination of visual cue assessment and in-context learning outperforms the baselines, even achieving a 100\% success rate at $\gamma=5$. The method without any feedback mechanism exhibits the lowest success rate and the highest localization errors. This is mainly caused by the hallucination of the VLM localizer that cannot be addressed in an open-loop manner. The baseline with only visual cue assessment checks the visual status of the first-round results to trigger the second-round inference, thus outperforming the baseline without any feedback mechanism. However, it fails to fully leverage the VLM checker's justification to guide the subsequent self-improvement in the right direction. In contrast, the full version of our closed-loop feedback mechanism utilizes the incorrect first-round results as negative examples for in-context learning, leading to more reasonable second-round implementations of the VLM localizer. More qualitative improvement results are provided in Supp. Mat., Sec.~C.

Open-loop LLM/VLM reasoning has inherent inference uncertainties due to the nature of probabilistic generative models~\cite{ye2025benchmarking,liu2025uncertainty}, which harms the system stability of interaction localization. To demonstrate that our closed-loop TIL paradigm holds low estimation uncertainties for the TIL task, we further present the standard deviation of each evaluation metric across 5 trials in Tab.~\ref{tab:uncertainty}. As can be seen, EgoLoc generates more stable estimation results than the other VLM-based baselines since it has the lowest standard deviations for all these metrics. This is primarily achieved by our proposed closed-loop feedback, which effectively reduces inference uncertainties by the second-round refinement. The experimental results also demonstrate that our method is better suited for downstream mixed reality and robotic applications, as these tasks typically require higher algorithmic stability to maintain effective long-term operation.

\textbf{VLM selection}. We also present EgoLoc performance with different VLMs on the EgoPAT3D-DT dataset, including GPT-4o~\cite{achiam2023gpt}, GPT-4o mini~\cite{achiam2023gpt}, Gemini 2.5 Pro~\cite{comanici2025gemini}, and Janus-Pro-7B~\cite{chen2025janus}. Tab.~\ref{tab:aba_vlm} shows that GPT-4o generates the best TIL estimations, which is the default VLM used in our proposed EgoLoc. Gemini 2.5 Pro presents the secondary performance, which remarkably outperforms the mini version of GPT-4o. 
While the open-source model, Janus-Pro-7B, exhibits relatively lower temporal localization accuracy, it offers practical advantages in enabling offline EgoLoc implementation and eliminating dependency on cloud-based API services. Besides, it still outperforms the baselines, including Threshold, VideoChat, and T-PIVOT ($N_\text{adj}=2$) shown in Tab.~\ref{tab:compare_main}. This also suggests that the systematic design in our EgoLoc paradigm can compensate for the performance degradation caused by the reduced parameter scale in VLM.

\subsection{Deployment on Downstream Tasks}
In this section, we deploy our proposed TIL paradigm on multiple downstream applications, which encompass robotic manipulation tasks in both simulated and real-world settings, showing a visual hint on the VR overlay for the incoming contact moment, and improving egocentric action recognition.

\subsubsection{Downstream Robotic Manipulation Tasks}
\label{sec:downstream_robot}
Here we first explore how our proposed TIL paradigm enables downstream robotic manipulation in the human-robot policy transfer framework.

\begin{figure}[t]
  \centering
  \captionsetup{aboveskip=2pt, belowskip=0pt}
  \includegraphics[width=1\linewidth]{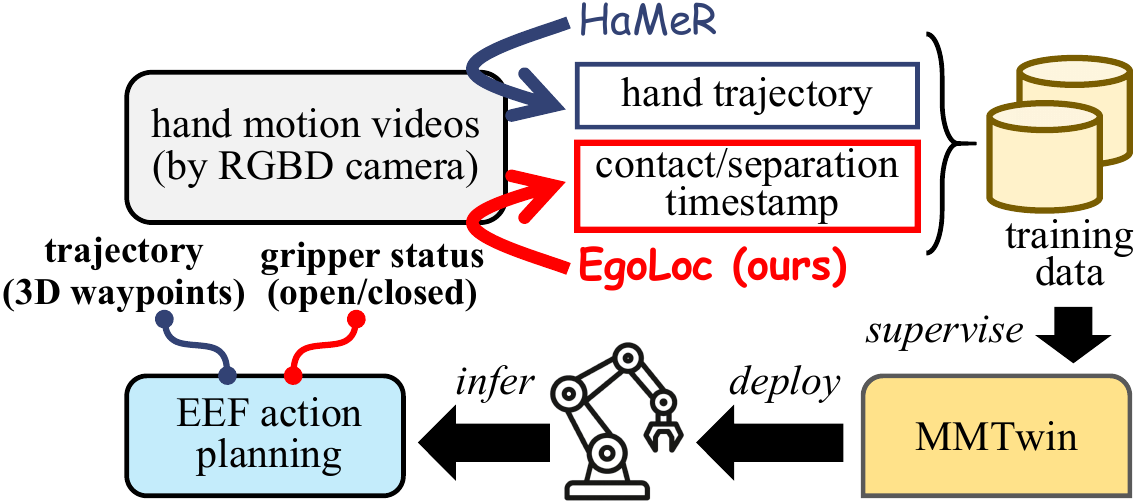}
  \caption{Illustration of the scheme to deploy EgoLoc on downstream robotic manipulation tasks.}
  \label{fig:robot_deploy_scheme}
  \vspace{-0.3cm}
\end{figure}

\textbf{{Deployment Scheme}}.
To demonstrate that interaction transitions captured by EgoLoc effectively guide human-robot action policy transfer, we modify the SOTA hand motion forecasting approach MMTwin~\cite{ma2025novel} to concurrently predict 3D hand trajectories and interaction transition timestamps. This is achieved by incorporating an additional decoder to convert denoised hand motion latents to the predicted hand-object contact and separation timestamps. The GT hand trajectories to supervise MMTwin are extracted by HaMeR~\cite{pavlakos2024reconstructing}, while our proposed EgoLoc automatically labels the GT hand-object contact and separation timestamps. Please refer to Supp. Mat., Sec.~B for more MMTwin training and inference details.
Afterwards, we directly use MMTwin optimized by the above labels to generate trajectories and grasp states for a robot end-effector (EEF) in the inference phase. We let the gripper follow the 3D waypoints of trajectory prediction, close at the predicted contact timestamp, and open at the predicted separation timestamp. The gripper orientation is determined by the heuristics of the prior work~\cite{papagiannis2024r}. The above deployment pipeline is illustrated in Fig.~\ref{fig:robot_deploy_scheme}. We implement the robotic manipulation tasks in both simulated and real-world settings with two different robot embodiments.

\begin{figure}[t]
  \centering
  \captionsetup{aboveskip=2pt, belowskip=0pt}
  \includegraphics[width=1\linewidth]{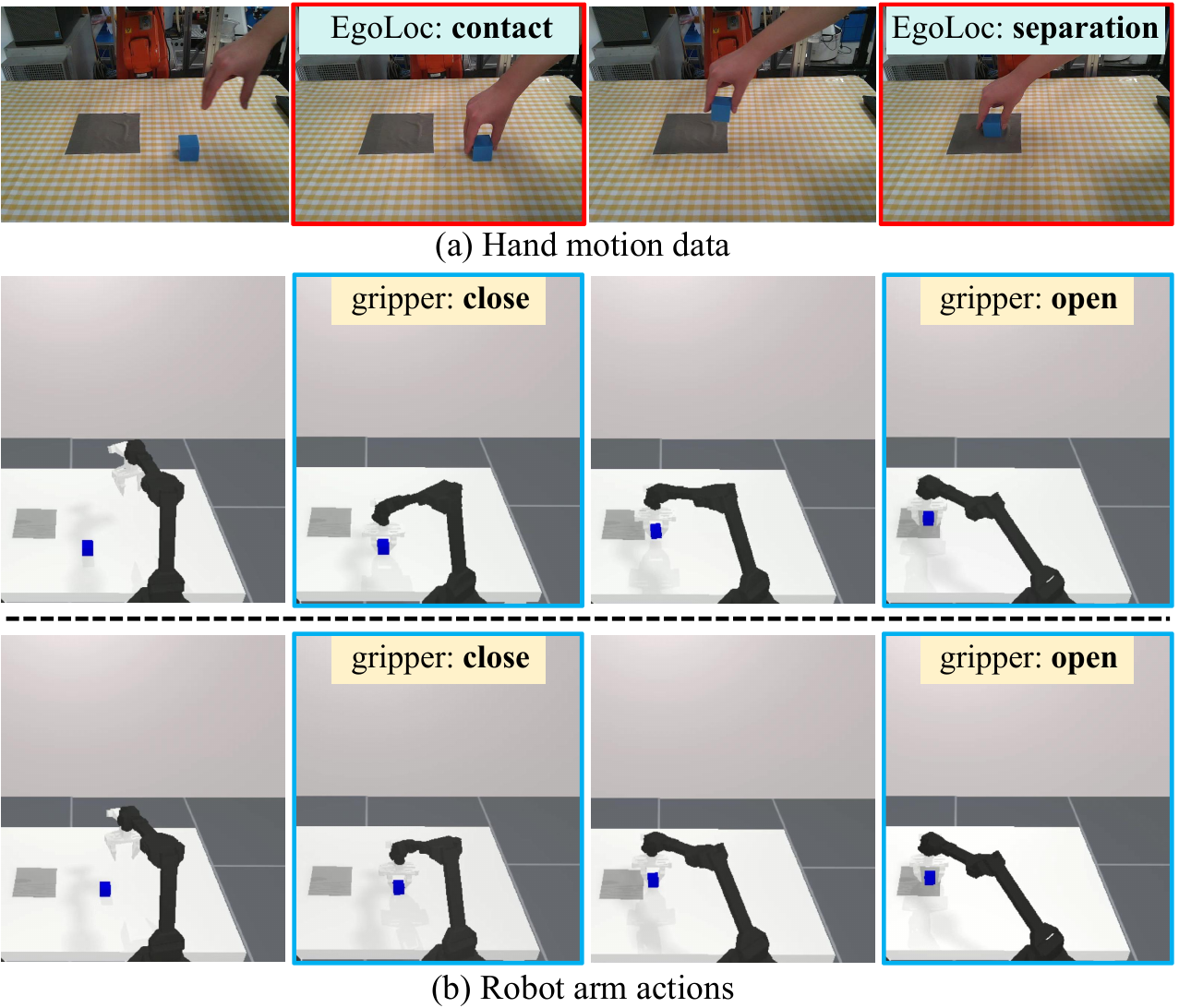}
  \caption{Visualization of collected hand motion data and transferred robot actions in the simulated environment. The red boxes in (a) denote the interaction transition timestamps localized by our EgoLoc, and the blue ones in (b) correspond to the gripper opening/closure moments offered by MMTwin~\cite{ma2025novel}.}
  \label{fig:robot_arm}
  \vspace{-0.3cm}
\end{figure}

\textbf{{Simulation Experiment}}.
We first efficiently validate the contribution of EgoLoc to the robotic pick-and-place task in a simulated environment.
Fig.~\ref{fig:robot_arm}(a) shows the scenario where we collect $400$ hand motion videos with a RealSense L515 camera for the task \textit{put the blue block onto the cloth}. The recorded data encompasses varying object positions to ensure reasonable optimization with diverse human movement patterns. We directly deploy the MMTwin model trained with the GT hand trajectories (from HaMeR) and interaction transition timestamps (from our EgoLoc) from these hand motion videos on a robot arm (ViperX 300) in the SAPIEN~\cite{xiang2020sapien} simulated environment. The visualized robot actions and quantitative results are presented in Fig.~\ref{fig:robot_arm}(b) and Tab.~\ref{tab:simulation_deploy}, respectively. 
We can see that the annotations of interaction transition timestamps from our EgoLoc effectively facilitate the robotic pick-and-place task by determining the timestamps of gripper opening/closing actions in the simulated environment. Tab.~\ref{tab:simulation_deploy} also demonstrates that our EgoLoc produces annotations closest to the manual labels, and significantly outperforms the baselines with higher success rates.

\begin{table}[t]
\small
\setlength{\tabcolsep}{4pt}
\center
\renewcommand\arraystretch{1}
\caption{Study on deployment on robotic manipulation tasks. We conduct 10 trials and record the number of successful attempts. Best results are viewed in \textbf{bold black}.}
\vspace{-0.1cm}
\begin{tabular}{l|cc}
\toprule
\multicolumn{1}{l|}{\multirow{1}{*}{Approach}}    & \makecell[c]{Simulated \\ (ViperX 300)}  & \makecell[c]{Real-world \\ (Franka Research 3)}    \\ \cmidrule{1-3} 
Manual (upper bound)  & \textit{9/10}     & \textit{10/10}  \\ \cmidrule{1-3} 
Threshold   & 6/10    &  2/10 \\
HOIMask   & 4/10    & 4/10 \\
Ours   & \textbf{8}/10   & \textbf{10}/10   \\
\bottomrule
\end{tabular}
\label{tab:simulation_deploy}
\vspace{-0.2cm}
\end{table}

\begin{figure*}[t]
  \centering
  \captionsetup{aboveskip=2pt, belowskip=0pt}
  \includegraphics[width=1\linewidth]{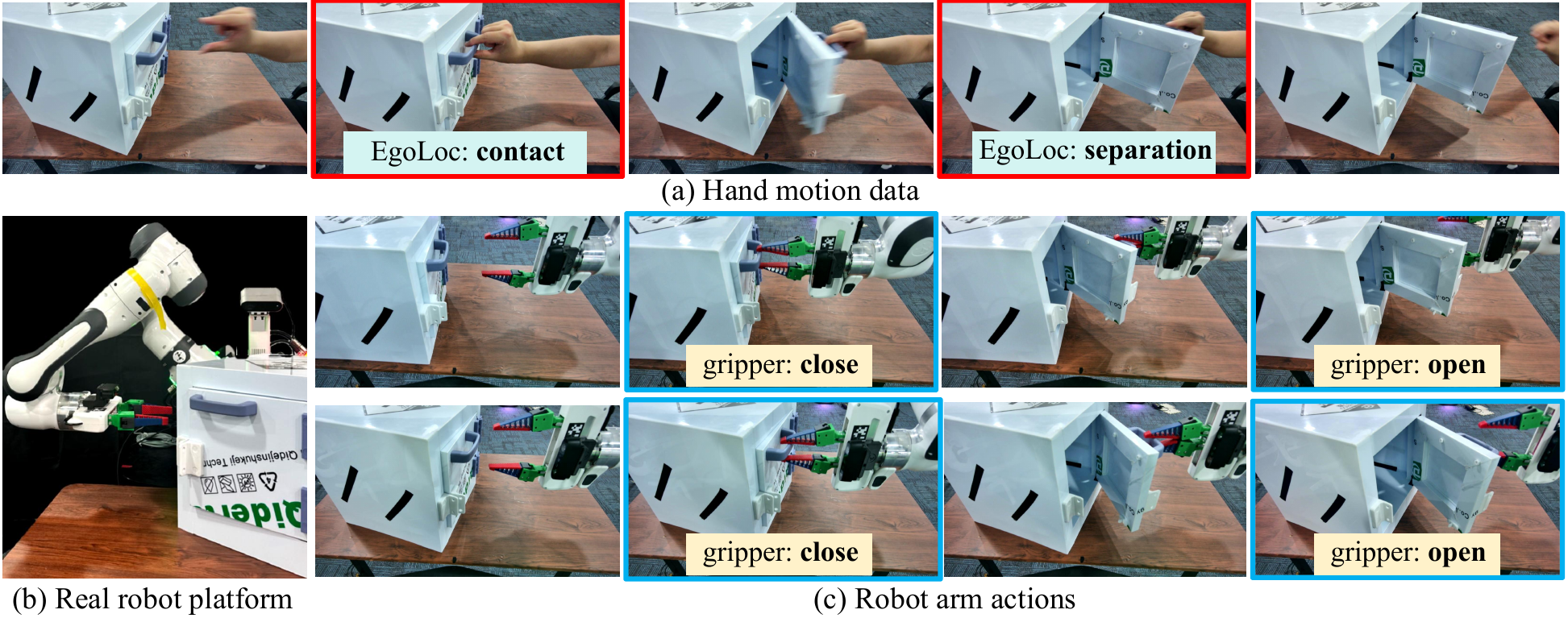}
  \caption{Visualization of collected hand motion data and transferred robot actions in the real-world environment. The red boxes in (a) denote the interaction transition timestamps localized by our EgoLoc, and the blue ones in (c) correspond to the gripper opening/closure moments offered by MMTwin~\cite{ma2025novel}.}
  \label{fig:robot_arm_real}
  \vspace{-0.3cm}
\end{figure*}

\textbf{{Real-World Experiment}}.
Moreover, we also validate the practicality of EgoLoc in a real-world robot platform shown in Fig.~\ref{fig:robot_arm_real}(b). Here, we let the Franka Emika Research 3 robotic arm with a parallel gripper perform the task \textit{open the cabinet door}, where a typical articulated object \textit{cabinet} is manipulated. We use an Orbbec Femto Bolt camera to record $400$ hand motion videos (see Fig.~\ref{fig:robot_arm_real}(a)), where GT trajectories and interaction transition timestamps are extracted by HaMeR and EgoLoc to train MMTwin as mentioned before. Fig.~\ref{fig:robot_arm_real}(c) shows that EgoLoc effectively guides robot action planning in real-world settings, and Tab.~\ref{tab:simulation_deploy} also presents that it results in higher success rates than the baselines. The experimental results demonstrate that our proposed EgoLoc facilitates downstream real-world robot tasks well by determining the gripper opening/closure moments. 

The findings in both simulated and real-world robotic manipulation tasks substantiate EgoLoc's capability to narrow the motion gap between different embodiments. It has the potential to be seamlessly integrated into modern frameworks of video-based human-robot skill transfer~\cite{ma2025novel,liu2025egozero,ren2025motion,anonymous2025articulated} in a generalizable zero-shot manner, which avoids costly tele-operation for policy learning.

\subsubsection{Downstream Egocentric Vision Tasks}
To showcase the broader applicability of our proposed TIL method, we further explore its potential in downstream egocentric vision tasks.

\begin{figure*}[t]
  \centering
  \captionsetup{aboveskip=2pt, belowskip=0pt}
  \includegraphics[width=1\linewidth]{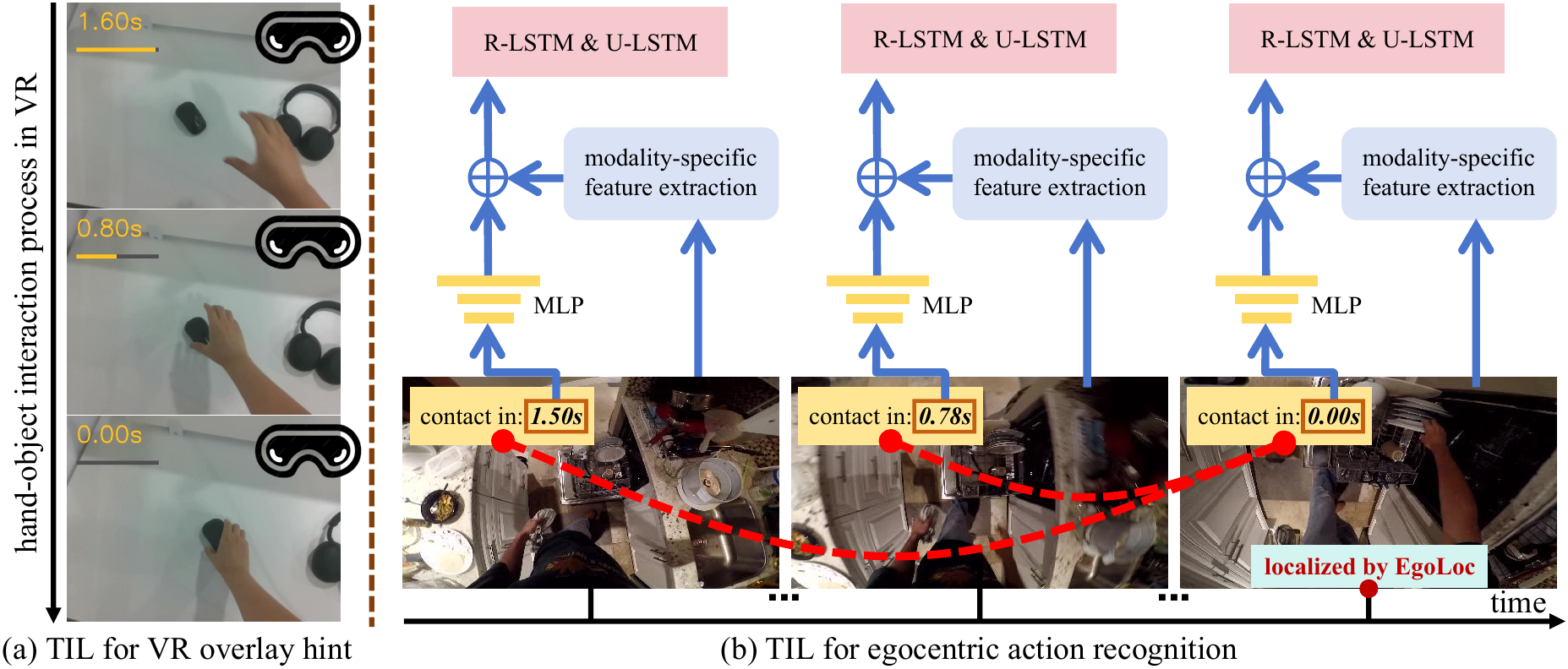}
  \caption{Illustration of how our proposed EgoLoc supports downstream egocentric vision tasks, including providing VR overlay hints for human activities and improving the existing action recognition framework.}
  \label{fig:downstream}
  \vspace{-0.3cm}
\end{figure*}

\textbf{VR Overlay Hint}. We simulate the role of EgoLoc in providing VR cues for human activities. We first train MMTwin with the interaction transition timestamps annotated by our EgoLoc as mentioned in Sec.~\ref{sec:downstream_robot}, on our proposed DeskTIL benchmark. In the deployment phase, we directly display potential time-to-contact for the next active object on the head-mounted camera's image output. As shown in Fig.~\ref{fig:downstream}(a), our TIL annotations help the prediction model capture future interaction transition timestamps, providing plausible VR cues for daily HOI activities. Thus, EgoLoc can annotate optimal contact timings in VR to train human motor skills (e.g., for chefs/athletes), issue preemptive warnings (e.g., ``Hot object contact in 2\,s''), or label predicted actions in collaborative VR (e.g., ``User A will pass an object in 2\,s'') to improve immersion and naturalness. 

\textbf{Egocentric Action Recognition}. Egocentric action recognition denotes the task of identifying human action categories based on visual input captured from a first-person perspective. Here we present how our proposed EgoLoc enhances the performance of the widely-used action recognition algorithm, RU-LSTM~\cite{furnari2020rolling}, on the Epic-Kitchens-55 dataset~\cite{damen2018scaling}. We split Epic-Kitchens-55 into training and test sets following the previous work~\cite{ma2024diff}. Fig.~\ref{fig:downstream}(b) shows how the TIL results from EgoLoc are integrated into RU-LSTM. Concretely, in addition to extracting modality-specific features for the egocentric image at each timestep in the input video clip, we transform the time-to-contact of each image to a feature vector with MLPs. Afterwards, we concatenate the modality-specific feature and time-to-contact feature as the input for the following LSTM-based rolling-unrolling operations. Due to cost considerations, we exploit Janus-Pro-7B instead of GPT-4o in EgoLoc for this experiment. Tab.~\ref{tab:action_recognition_deploy} shows that the interaction transition timestamps localized by EgoLoc enhance RU-LSTM's action recognition performance. This is achieved by strategically injecting richer human-object interaction transition information, enabling more precise modeling of human motion patterns in egocentric views.

\begin{table}[t]
\small
\setlength{\tabcolsep}{10pt}
\center
\renewcommand\arraystretch{1}
\caption{Study on enhancement for action recognition. We report Top-5 action accuracy (\%) of RGB and flow branches in RU-LSTM. Best results are viewed in \textbf{bold black}.}
\vspace{-0.1cm}
\begin{tabular}{l|cc}
\toprule
\multicolumn{1}{l|}{\multirow{1}{*}{Approach}}    & RGB branch  & Flow branch    \\ \cmidrule{1-3} 
Vanilla  & 41.78     & 30.96 \\ 
Enhanced by EgoLoc  & \textbf{42.20}  & \textbf{31.22} \\
\bottomrule
\end{tabular}
\label{tab:action_recognition_deploy}
\vspace{-0.2cm}
\end{table}

These experimental results on downstream egocentric vision tasks reveal that interaction transition timestamps embody essential patterns of human activities. Temporal interaction localization enables deeper mining of these human motion characteristics, and conversely facilitates enhanced guidance of human behaviors.

\vspace{-0.1cm}
\section{Conclusion}
\vspace{-0.1cm}
\label{sec:conclusion}
In this paper, we propose EgoLoc, a novel temporal interaction localization method for egocentric videos. It offers a generalizable solution to accurately localize the timestamps of hand-object contact and separation in a zero-shot manner. 
We design a hand-dynamics-guided sampling strategy to produce high-quality initial guesses for VLM-based localization, coupled with a closed-loop feedback mechanism to further refine the TIL results. 
The experimental results on the publicly available dataset and our newly proposed benchmarks demonstrate that EgoLoc achieves plausible temporal interaction localization on egocentric videos featuring different lengths and HOI counts. It generalizes well across multiple scenes, and has the potential to enable multiple downstream applications. We hope that our proposed TIL paradigm and benchmarks will facilitate future work in the literature. We also expect that TIL can foster an increasing range of downstream applications in mixed reality and robot policy generation. In future work, we will explore TIL for bimanual HOI transitions and rapid actions.

\small
\bibliographystyle{ieeetr}
\bibliography{main}

\newpage
\setcounter{section}{0}
\setcounter{figure}{0}
\setcounter{table}{0}
\renewcommand{\thesection}{\Alph{section}}
\renewcommand{\thefigure}{\Alph{figure}}
\renewcommand{\thetable}{\Alph{table}}

\begin{flushleft}
    \huge Supplementary Material
\end{flushleft}

\section{Statistical Properties of DeskTIL and ManiTIL}

\begin{figure}[htbp]
  \centering
  \begin{subfigure}[b]{0.85\linewidth}
  \includegraphics[width=1\linewidth]{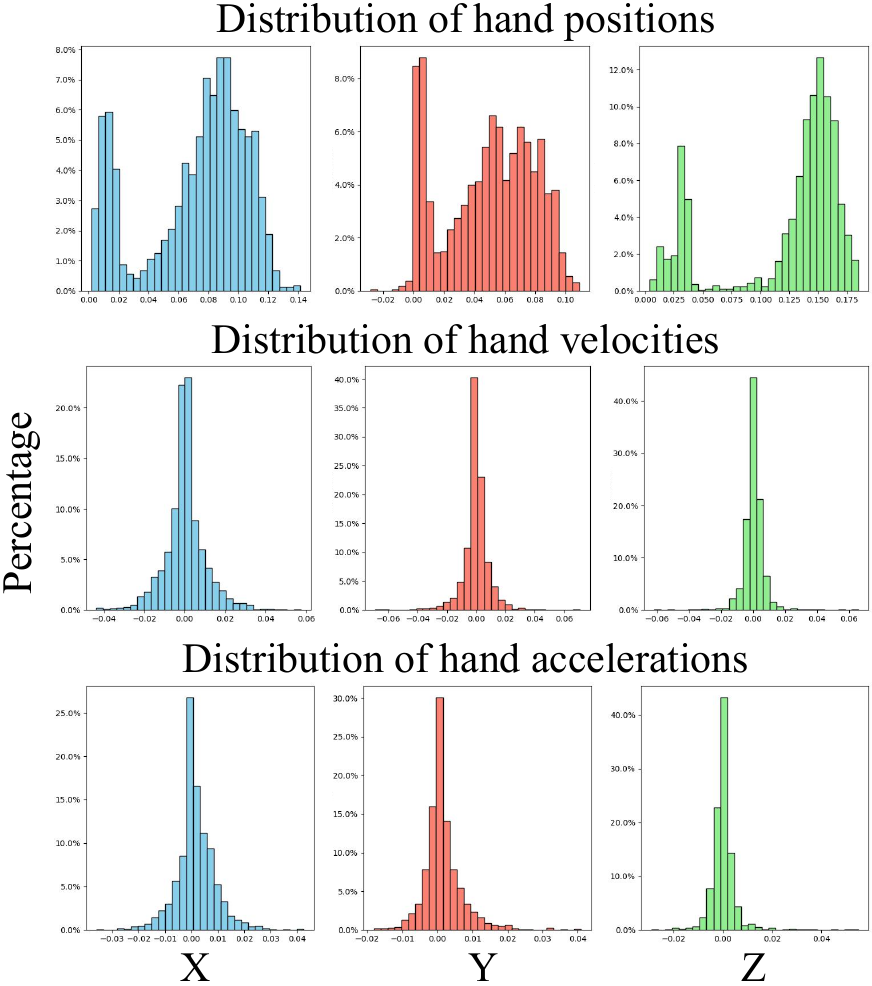}
    \caption{DeskTIL}
    \label{fig:sub1}
  \end{subfigure}
  \begin{subfigure}[b]{0.85\linewidth}
  \includegraphics[width=1\linewidth]{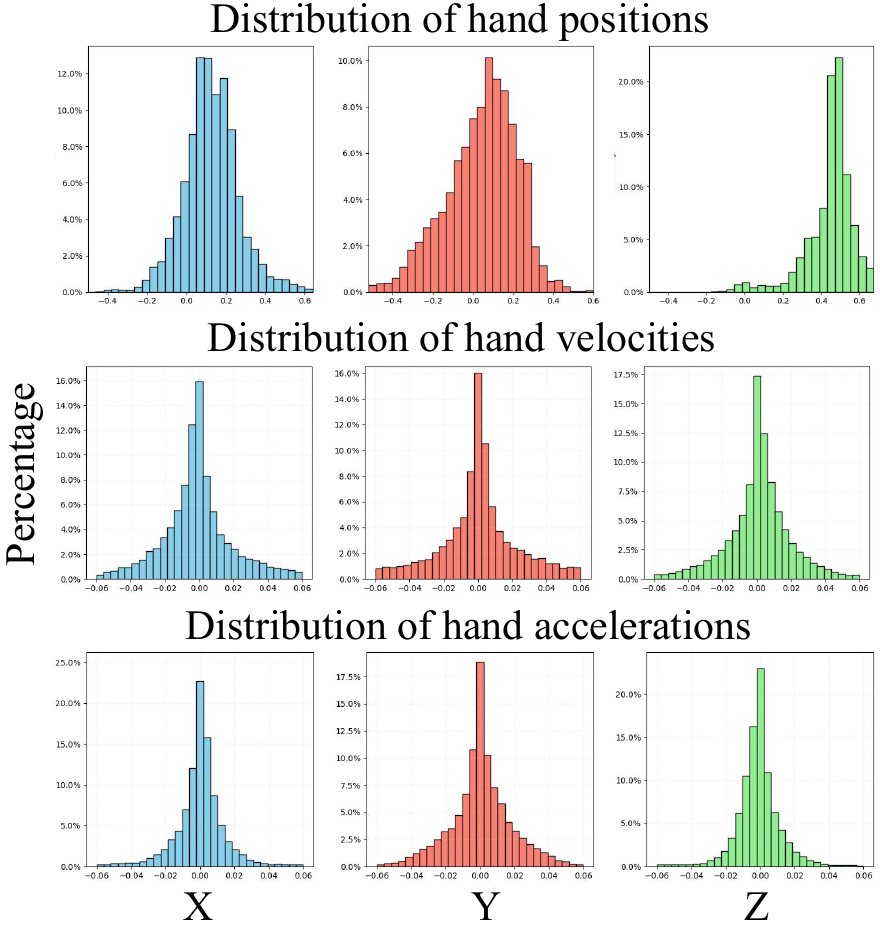}
    \caption{ManiTIL}
    \label{fig:sub2}
  \end{subfigure}
  \vspace{-0.3cm}
  \caption{Statistical characterization of hand motion patterns across DeskTIL and ManiTIL.}
  \label{fig:statistics_desktil_manitil}
\end{figure}

In this section, we present a comprehensive statistical analysis of hand motion characteristics in our newly proposed DeskTIL and ManiTIL benchmarks. We hope these statistical results can serve as foundational references for future temporal interaction localization (TIL) studies regarding diverse human hand motion patterns. Fig.~\ref{fig:statistics_desktil_manitil} shows the distributions of 3D hand positions, velocities, and accelerations in the xyz coordinate. As can be noted, the hand movements in our benchmarks exhibit natural spatial patterns, with higher concentration zones near typical workspace areas and more dispersed distributions in peripheral regions. The high proportion of zero-velocity frames in the hand motion data provides empirical evidence that our benchmarks include extensive hand-object interaction phases. Additionally, our collected data encompasses the extensive z-axis hand movements that are commonly observed in daily-life activities like reaching for objects. Such depth variations can cause severe scale ambiguity in 2D interaction localization. As introduced in the main text, our proposed EgoLoc overcomes this challenge by incorporating 3D perception into hand dynamics analysis.

\section{MMTwin Training and Inference Details}

Sec.~4.4 of the main text has introduced how our proposed EgoLoc facilitates downstream applications by enriching the output of the SOTA hand motion forecasting approach MMTwin~\cite{ma2025novel}. EgoLoc automatically annotates hand-object contact and separation states to generate training labels. After label generation, we train MMTwin using AdamW optimizer~\cite{kingma2014adam} with a learning rate of 5e-5 for 30K epochs on our self-recorded hand motion videos for each downstream task. All the modules are trained with a batch size of $32$ on $2$ A100 GPUs. We integrate $6$ egomotion-aware Mamba (EAM) blocks and $1$ structure-aware Transformer block in the denoising model. The additional decoder, which converts denoised hand motion latents to predicted hand-object contact and separation timestamps, is set to the same MLPs as the trajectory decoder in MMTwin, except for the output dimension $2$. The total number of diffusion steps for MMTwin inference is set as $1000$. The other model parameters follow those specified in the original paper~\cite{ma2025novel}.

\begin{figure}[t]
  \centering
  \captionsetup{aboveskip=2pt, belowskip=0pt}
  \includegraphics[width=1\linewidth]{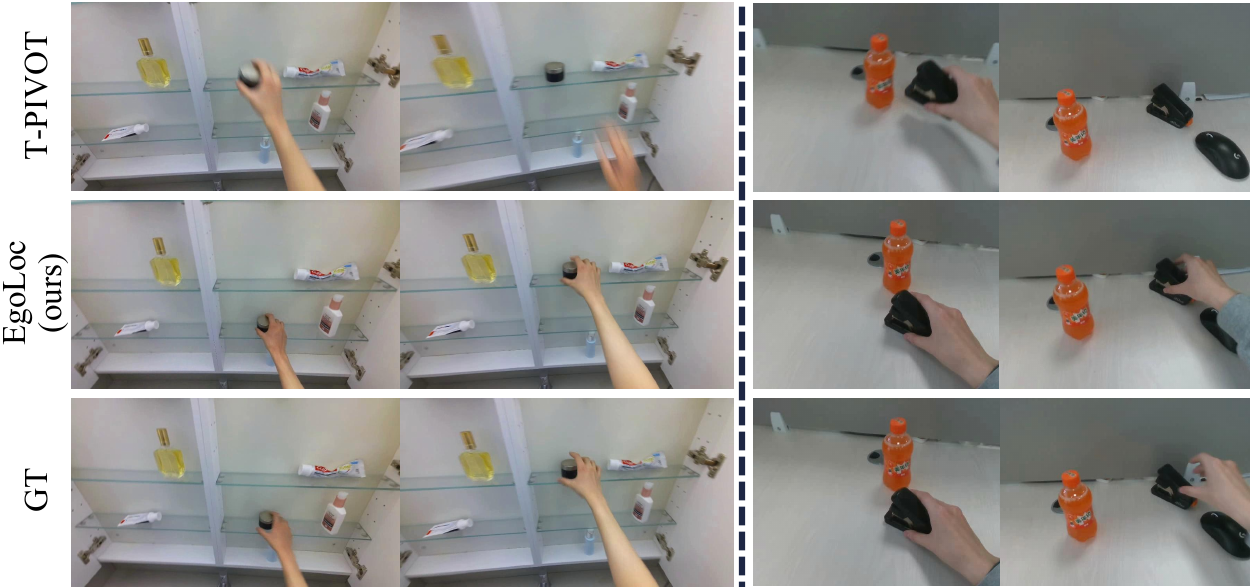}
  \caption{Visualization of TIL results from EgoLoc, T-PIVOT, and GT annotations (left: EgoPAT3D-DT, right: DeskTIL).}
  \label{fig:results_supp}
\end{figure}

\begin{figure*}[t]
  \centering
  \captionsetup{aboveskip=2pt, belowskip=0pt}
  \includegraphics[width=0.6\linewidth]{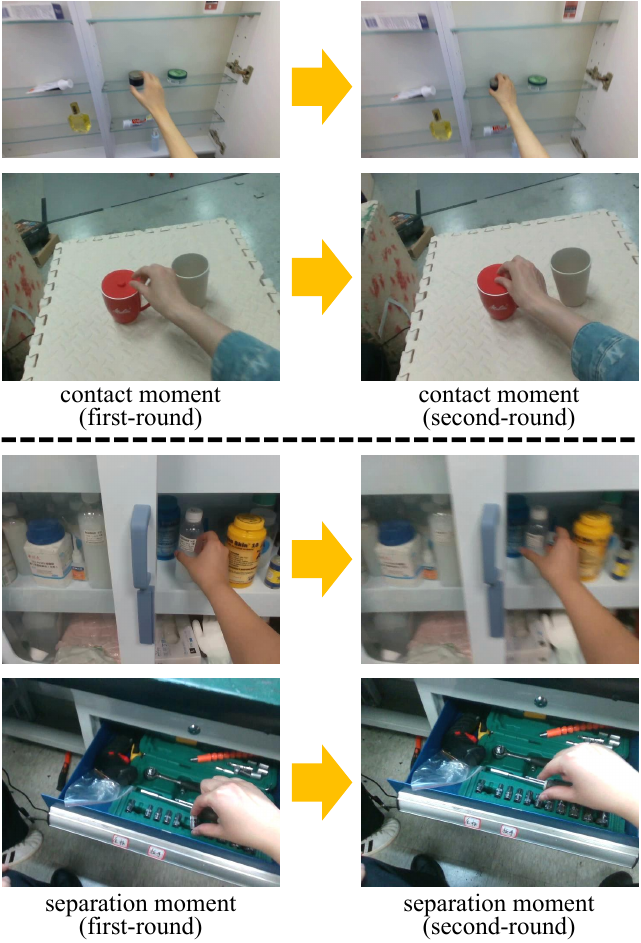}
  \caption{Visualization of the refinement from our proposed closed-loop feedback mechanism.}
  \label{fig:refinement}
  \vspace{-0.5cm}
\end{figure*}

\section{More Visualization Results}
We have provided multiple visualizations of TIL results from EgoLoc and baseline methods in the main text. In Fig.~\ref{fig:results_supp}, we further illustrate the qualitative comparison between our proposed EgoLoc and T-PIVOT~\cite{wake2024open} (both with $N_\text{adj}=2$). As can be seen, our proposed EgoLoc extracts more plausible interaction transition timestamps, thanks to the devised sampling strategy and closed-loop feedback mechanism. Note that we do not test T-PIVOT on ManiTIL since it is not trivial to adapt it to localize timestamps for the video with multiple HOI events.

In addition, we further visualize the refinement of localization results by our proposed closed-loop feedback mechanism in EgoLoc. As shown in Fig.~\ref{fig:refinement}, the incorrect first-round TIL results (left) rejected by the VLM checker are utilized as negative examples for in-context learning, which facilitates the VLM localizer to produce more precise second-round results (right).

\end{document}